\documentclass[10pt]{article}

\usepackage[a4paper,margin=1in]{geometry}

\usepackage[T1]{fontenc}
\usepackage{lmodern}

\usepackage[round]{natbib}

\usepackage{float}
\usepackage{amssymb}
\usepackage{booktabs}
\usepackage{subcaption}

\usepackage{graphicx}
\usepackage{algorithm}
\usepackage{algpseudocode}

\usepackage{mathtools}
\mathtoolsset{showonlyrefs}
\usepackage{enumitem}

\usepackage{hyperref}
\usepackage{cleveref}

\title{Sparse Bayesian Neural Networks via Sticky Piecewise Deterministic Markov Processes}

\usepackage{authblk}

\author[1,*]{August Arnstad}
\author[1]{Geir Olve Storvik}
\author[2,*]{Filippo Pagani}
\author[1,*]{Leiv R{\o}nneberg}
\affil[1]{Department of Mathematics, University of Oslo, NO}
\affil[2]{Department of Statistics, University of Warwick, UK}

\date{}

\begin{document}

\maketitle

{\renewcommand\thefootnote{*}%
\begin{NoHyper}\footnotetext{Corresponding authors: \texttt{augusa@math.uio.no}, \texttt{filippo.pagani@warwick.ac.uk}, \texttt{ltronneb@math.uio.no}}\end{NoHyper}}

\begin{abstract}
    There is growing evidence that the performance of a neural network depends largely on a small subset of its weights, which are typically identified by training a dense network and pruning it afterwards. Within a Bayesian framework, pruning can be directly incorporated through a spike-and-slab prior, providing a proper treatment of the uncertainties involved, both with respect to pruning and to parameter uncertainty.
    In Bayesian neural networks (BNNs), sampling by MCMC is notoriously hard. We demonstrate that modern Monte Carlo methods based on piecewise deterministic Markov processes (PDMPs) can be used to sample posteriors of BNNs asymptotically exact whenever numerically constructed rate bounds hold. Spike-and-slab priors can be incorporated through  \emph{sticky} versions of the PDMPs, setting weights exactly to zero during sampling. We examine the predictive performance and convergence of the samplers, and show that sparse BNNs sampled this way can be more robust to input noise than dense ones.
\end{abstract}

\section{Introduction}
Bayesian neural networks (BNNs) quantify uncertainty for models that are otherwise prone to overconfidence, but characterizing their posterior over the network weights is difficult. Variational inference \citep{vi_2011_graves} replaces the posterior with a tractable distribution, whereas stochastic gradient samplers \citep{Welling2011BayesianLV,sghmc_2014_chen} use minibatches, which scales well but introduces bias. Asymptotically exact samplers target the posterior without bias and have been shown feasible for larger networks \citep{izmailov_hmc_bnn, sommer2024_large_BNNs}, but remain too costly to be practical.

\begin{figure}[t]
\centerline{\includegraphics[width=0.6\textwidth]{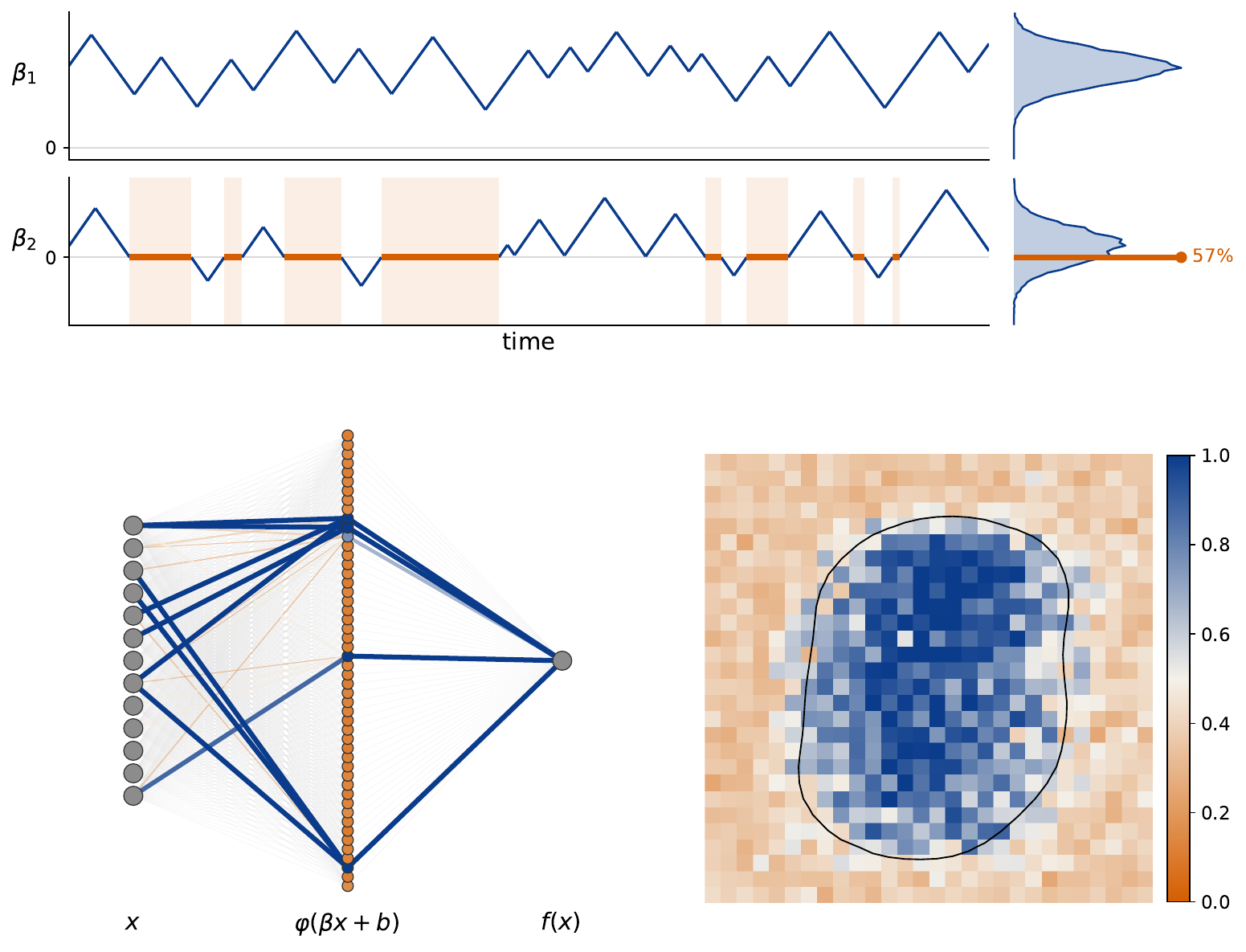}}
\caption{
Sticky PDMPs sample exact zeros. \textbf{(a)} Example of a sticky ZigZag parameter path. \textbf{(b)} Posterior inclusion probabilities of weights in a BNN sampled with sticky PDMPs. \textbf{(c)} Inclusion probabilities of the MNIST input pixel weights, low where pixels carry no information.
}
\label{fig:headliner}
\end{figure}

Computational cost is not the only problem in BNNs. The networks are highly over-parametrized, which makes Bayesian inference difficult as the parameters are not identifiable due to symmetries \citep{papamarkou2022challenges}. Furthermore, weights that do not affect the data fit, such as those connected to inputs that are zero across the training set, retain their prior variance in the posterior, which can make BNNs less robust to covariate shift than networks trained by SGD \citep{izmailov_hmc_bnn, izmailov2021dangers}. Setting such weights to zero removes this source of variance, and BNNs can indeed be pruned substantially with little loss of predictive performance \citep{molchanov2017variational, hubin_sparse_bnn}, suggesting that a subset of the parameters suffices to characterize the posterior predictive. The canonical method for sparsifying a Bayesian model is the spike-and-slab prior \citep{mitchell1988bayesian}, which places positive prior probability on parameters being exactly zero, so that parameters the data do not inform are exactly zero in most draws. Traditionally, it has been hard to make use of spike-and-slab priors in Bayesian deep learning, as the efficient gradient-based samplers require a differentiable density, which a prior with point masses at zero does not have. Thus, such priors have been applied only approximately \citep{hubin_sparse_bnn}.

Piecewise deterministic Markov process samplers~\citep[PDMPs,][]{fearnhead2018piecewise} are a class of non-reversible, continuous-time Markov chain Monte Carlo (MCMC) samplers. A trajectory is simulated over a time horizon $T$ from a process that evolves deterministically between random event times, constructed so that the target distribution is the position marginal of its stationary distribution. Once the trajectory is long enough, samples are obtained by evaluating the position at time points along the path. Such methods have received great interest recently \citep{bierkens2019bigdata, corbella2022, fearnhead2024stochasticgradientpiecewisedeterministic, chevallier2025}, as exact subsampling can keep their cost per effective sample constant in the dataset size (\textit{super-efficiency}). Furthermore, the PDMP samplers can be made \textit{sticky} \citep{bierkens2022sticky}, so that they target the posterior under a spike-and-slab prior directly, and sparsity arises during sampling without any post-processing or thresholding.

\paragraph{Contributions}
In this paper, we show that PDMPs can sample the posterior of a BNN under a spike-and-slab prior, which mixes a continuous density with a point mass at zero. With full-batch gradients, the sampling is asymptotically exact, provided the numerical bounds used to simulate the event times hold. This brings PDMPs closer to practical use in Bayesian deep learning. In particular, we
\begin{itemize}[nosep]
    \item adapt sticky and non-sticky PDMPs to BNNs with cheap numerical bounds;
    \item provide a PyTorch implementation for any network module with GPU acceleration;
    \item evaluate the samplers against NUTS and a sparse variational method;
    \item show that exact zeros give high sparsity and can improve robustness to input noise.
\end{itemize}

\section{Related work}
The Bayesian perspective on neural networks was introduced in \citet{mackay_bayesian_1992}, with Hamiltonian Monte Carlo (HMC) popularized for Bayesian inference in \citet{neal_bayesian_1996}. HMC remains the gold standard for asymptotically exact sampling, but the scale of modern neural networks has made faster, approximate techniques such as variational inference (VI) attractive \citep{vi_2011_graves, blundell_vi}.

Regardless of sampling technique, the choice of prior has always been important. While the i.i.d.\ Gaussian prior has been the standard choice \citep{fortuin2022priors}, it has also been subject to criticism \citep{wenzel2021bayesianneuralnetworkposteriors, tran_all_2022}. This has sparked literature on priors that are better designed to handle neural networks. \citet{fortuin2021bayesian} find that the weights of networks trained by SGD are heavy-tailed, motivating priors such as the Laplace and Student-t, but these do not resolve the lack of robustness of BNNs under covariate shift \citep{izmailov2021dangers}. Sparsity-inducing global-local priors such as the horseshoe instead shrink most weights towards zero \citep{louizos_bayesian_compression_2017, ghosh2019model}, but rely on specifying a threshold for pruning after sampling, which can be challenging.
The spike-and-slab prior instead assigns a non-zero probability to a parameter being exactly zero. \citet{Polson_posterior_concentration} establish posterior concentration at the near-minimax rate for deep networks under such a prior, but propose no algorithm for sampling the posterior they describe. Within VI, where the atom can be placed in the variational family by construction, \citet{hubin_sparse_bnn} report substantial sparsification at little cost to predictive accuracy.

\textbf{PDMP} samplers originate in the physics literature \citep{Peters_de_With_2012_phsyics_PDMP}, and were brought to general use with the bouncy particle sampler (BPS) of \citet{bouchard2017}, followed by the ZigZag \citep{bierkens2019bigdata} and Boomerang \citep{bierkens2020Boomerang} samplers.

Implementing PDMPs is non-trivial, as event times are drawn from an inhomogeneous Poisson process, usually by thinning \citep{lewis_shedler_poisson_thinning}, which requires an upper bound on the rate. \citet{sutton2023concave} derive bounds from a concave-convex decomposition of the rate when it has a closed form. For general targets, \citet{corbella2022} construct the bounds numerically on short horizons, and \citet{andral2024} make this adaptive with a tighter, grid-based bound implemented in JAX \citep{jax2018github}. \citet{goan2023piecewise} apply PDMPs to BNNs in TensorFlow \citep{tensorflow2015-whitepaper}, with a different bounding strategy. \citet{chevallier2025} instead discretise the dynamics and correct the discretisation error with a Metropolis adjustment, removing the need for explicit bounds.

Sticky PDMPs were introduced by \citet{bierkens2022sticky} to sample targets that mix a continuous density with point masses at zero, such as spike-and-slab posteriors. Parameters can then stay exactly zero, giving non-reversible moves between sub-models \citep{bierkens2022sticky}, closely related to the reversible jump PDMPs of \citet{Chevallier02102023}. Later, stickiness was generalized as a boundary condition, alongside hard and soft walls, and teleportation \citep{bierkens2023methodsapplicationspdmpsamplers}. \citet{fearnhead2024stochasticgradientpiecewisedeterministic} introduced stochastic gradient PDMPs, including a sticky ZigZag, which discretise time to avoid rate bounds at the cost of a bias of the order of the step size, and applied them to a BNN with one hidden layer. To the best of our knowledge, continuous-time sticky PDMPs have only been applied to targets with a tractable upper bound on the rate.

\section{Bayesian neural networks}
\label{sec:bnns}
For some input $\mathbf{x}$, let $f(\mathbf{x}; \boldsymbol{\beta})$ denote a neural network with parameters $\boldsymbol{\beta}\in\mathbb{R}^d$ consisting of weights and biases. We seek the posterior distribution $p(\boldsymbol{\beta}|\mathcal{D}) \propto p(\mathcal{D}| \boldsymbol{\beta})p(\boldsymbol{\beta})$ where $\mathcal{D}$ denotes the training data $\{(\mathbf{x}_i, \mathbf{y}_i)\}_{i=1}^n$, $p(\mathcal{D}| \boldsymbol{\beta})$ is the likelihood and $p(\boldsymbol{\beta})$ the prior. Predictive distributions for unseen data $(x, y)$ can be calculated as
\begin{equation}
    \label{eq:model}
    p(y | x, \mathcal{D}) = \int p(y | x, \boldsymbol{\beta}) p(\boldsymbol{\beta} | \mathcal{D})d\boldsymbol{\beta} \ ,
\end{equation}
where $p(y | x, \boldsymbol{\beta})$ is the predictive distribution of the data for some value of the parameters. This integral is not available in closed form for neural networks. However, if samples $\boldsymbol{\beta}^1,...,\boldsymbol{\beta}^S$ from the posterior $p(\boldsymbol{\beta}|\mathcal{D})$ are available, a Monte Carlo estimate of the predictive distribution can be applied:
\begin{equation}
\label{eq:pred}
    p(y | x, \mathcal{D}) \approx \tfrac{1}{S}\sum_{s=1}^S p(y | x, \boldsymbol{\beta}^s).
\end{equation}
Posterior samples are typically drawn with discrete-time MCMC such as HMC. We will consider the alternative PDMPs, which are continuous time samplers and extend naturally to spike-and-slab priors.

\section{Piecewise deterministic Markov processes}
\label{sec:pdmps}

A piecewise deterministic Markov process (PDMP) evolves deterministically between random event times, at which its state changes stochastically. PDMP samplers target an extended distribution $\pi(\boldsymbol{\beta},\boldsymbol{v})$ with $p(\boldsymbol{\beta}|\mathcal{D})$ as its marginal, where the auxiliary \textit{velocity} $\boldsymbol{v}$ drives the deterministic motion. With $\boldsymbol{z}(t)=(\boldsymbol{\beta}(t),\boldsymbol{v}(t))$, a PDMP is specified by
\begin{itemize}[nosep]
    \item a deterministic flow between events, given as the solution of an ordinary differential equation $\Phi_t$;
    \item an event rate $\lambda(\boldsymbol{z}(t))$ determining when events occur;
    \item a transition kernel $Q(\cdot | \boldsymbol{z}(t))$ describing how the state changes at an event.
\end{itemize}
A simple example is the \emph{ZigZag} sampler~\citep{bierkens2019bigdata}, with $\boldsymbol{v}\in\{-1,1\}^d$ and linear motion
\begin{equation}\label{eq:zigzag}
    \Phi_t(\boldsymbol{\beta}, \boldsymbol{v}) = (\boldsymbol{\beta} + t\boldsymbol{v},\ \boldsymbol{v}) \ .
\end{equation}

The time to the next event $\tau$ is the first event of an inhomogeneous Poisson process with rate function $\lambda(t)$.
The function $\lambda(t)$ is defined through the distribution $\pi(\boldsymbol{\beta},\boldsymbol{v})$. In particular, by defining
\begin{equation}
\label{eq:energy_func}
U(\boldsymbol{\beta}) = -\log p(\mathcal{D}| \boldsymbol{\beta}) - \log p(\boldsymbol{\beta}),
\end{equation}
we have for the ZigZag
\begin{align*}
    \lambda(t) =& \sum_{i=1}^{d} \lambda_i(t) , \\
\lambda_i(t) = &\max\{v_i(t)\tfrac{\partial}{\partial\beta_i} U(\boldsymbol{\beta}(t)), 0\} \quad i=1, \dots d.
\end{align*}
Then, the parameter $\beta_i$ where the change occurs is selected with probability $\lambda_i(\tau)/\lambda(\tau)$. Upon an event in parameter $\beta_i$, the transition kernel of the ZigZag reverses the velocity using the flip operator
\begin{equation}
    F_i(\boldsymbol{\beta}, \boldsymbol{v}) = (\boldsymbol{\beta}, \boldsymbol{v}'),
    \qquad
    v'_j = \begin{cases} -v_j & j = i \\ v_j & j \neq i \, . \end{cases}
\end{equation}

The main computational challenge for PDMP samplers is the simulation of the next event time. Along the deterministic trajectory, the event rate generally varies with time. In general, assuming current time is zero, the next event has survival function
\begin{equation}
\label{eq:event_time}
    \mathbb{P}(\tau>t) = \exp\left(-\int_0^t \lambda(s) \, \mathrm{d}s \right).
\end{equation}
If the integral $\Lambda(t)=\int_0^t\lambda(s)ds$ can be inverted in closed form (e.g. constant or piecewise constant $\lambda$), $\tau$ can be sampled by inversion, setting $\tau = \Lambda^{-1}(E)$ with $E\sim\mathrm{Exp}(1)$.

As the flow is deterministic, the process is fully determined by its \emph{skeleton points} $(T_k, B_k, V_k)_{k\geq0}$, the event times $T_k$, with $T_0=0$, and the states $B_k=\boldsymbol{\beta}(T_k)$ and $V_k=\boldsymbol{v}(T_k)$, which form a discrete-time Markov chain \citep{Bierkens_limits_2017}. Skeleton points are not samples from the target, as they concentrate where the rate is high, in its tails \citep{bierkens2019bigdata}. Samples are instead read off the trajectory on $[0,T]$, for example at equally spaced times $t_s=sT/S$ in~\eqref{eq:pred}, or at uniformly drawn times as we do (\Cref{sec:supp_samples}).

An alternative to the ZigZag sampler is the Boomerang sampler~\citep{bierkens2020Boomerang}, which moves on elliptical paths around a Gaussian reference measure chosen to resemble the posterior. We also consider the Boomerang sampler, and describe it in \Cref{sec:supp_Boomerang}.

\subsection{Sticky PDMPs}
\label{sec:sticky_pdmps}
The ZigZag and Boomerang samplers target distributions with a density on $\mathbb{R}^d$, meaning all coordinates of $\boldsymbol{\beta}$ are almost surely nonzero. However, sparse networks, where some of the weights are put to zero, can have both statistical and computational benefits. A particularly interesting prior is the spike-and-slab prior
\begin{equation}
p(d\boldsymbol{\beta})
=\prod_{i=1}^d(w_ip_i({\beta_i})d\beta_i+(1-w_i)\delta_0(d\beta_i))
\end{equation}
where $p_i$ is the slab density, $w_i\in[0,1]$ and $\delta_0$ is the Dirac measure at zero. \citet{bierkens2022sticky} introduced the \emph{Sticky} PDMP samplers for such settings.

The notion of stickiness is introduced as a new event type, independent of the rate function. If the trajectory of parameter $\beta_i$ hits zero, its velocity is set to zero, freezing the parameter for an exponential time with rate $\lvert v_i \rvert \kappa_i$, where $v_i$ is the velocity as the parameter hits $0$ while $\kappa_i=w_ip_i(0)/(1-w_i)$. After staying frozen for the exponentially distributed time, the parameter thaws and continues motion with the same velocity it had before freezing. Stickiness thus allows parameters to spend a positive amount of time exactly at zero, and the fraction of time a parameter is frozen estimates its posterior probability of being zero \citep{bierkens2022sticky}.

In practice, we treat freezing and thawing as separate events. Let $A$ and $F$ denote the active and frozen coordinates, and $\lambda_A(t)$ the reflection rate summed over $A$ (since $\lambda_i(t)=0$ for $i\in F$). Freezing times are deterministic functions of the trajectory, and we care only about the next active coordinate that hits zero. Upon freezing, coordinate $i$ draws a thawing clock $s_i \sim \mathrm{Exp}(|v_i|\kappa_i)$, with $|v_i|=1$ for the ZigZag, which counts down as time elapses.
For more details on the construction of stickiness we refer to \citet{bierkens2022sticky} and one can see \citet{schauer2021zigzagboomerang} for implementations.

\subsection{Piecewise constant bounds}
\label{sec:bounds}
As the solution to the integral in \eqref{eq:event_time} is generally not available in closed form, practical implementations often rely on Poisson thinning~\citep{corbella2022, bertazzi2022, sutton2023concave}. Poisson thinning~\citep{lewis_shedler_poisson_thinning} proposes an event time $\tau$ based on a tractable upper bound $\tilde{\lambda}(t) \geq \lambda(t)$ and accepts it with probability $\lambda(\tau)/\tilde{\lambda}(\tau)$. Following \citet{corbella2022}, the rate is only bounded over a short window $[0, t_{\max}]$ from the current state. The process is moved deterministically to $\min\{\tau,t_{\max}\}$ and a new window starts from there. If the event is accepted, a new skeleton point appears at $\tau$. For the sticky samplers, the window is cut short at the horizon $h$, the earliest of $t_{\max}$, the next freezing time and the thawing clocks, and reflections are proposed by thinning on $[0,h]$, see~\Cref{alg:sticky}.

\begin{algorithm}[t]
\caption{Sticky ZigZag step. Every reflection, freeze and thaw is a skeleton point.}\label{alg:sticky}
\begin{algorithmic}[1]
\Require $(\boldsymbol{\beta}, v)$,  $t_{\max}$, active set $A$, frozen set $F$ with clocks $s_i$
\State $h \gets \min\big(t_{\max},\ \min_{i \in A,\, \beta_i v_i < 0} \lvert \beta_i \rvert,\ \min_{i \in F} s_i\big)$
\State build $\tilde\lambda \geq \lambda_A(t)$ on $[0, h]$, $u \gets 0$
\Repeat
    \State $u \gets u + e$, \ $e \sim \mathrm{Exp}(\tilde\lambda)$
\Until{$u \geq h$ or $\omega < \lambda_A(u)/\tilde\lambda$, \ $\omega \sim \mathcal{U}(0, 1)$}
\State $\tau \gets \min(u, h)$, \ $\boldsymbol{\beta} \gets \boldsymbol{\beta} + \tau v$, \ $s_i \gets s_i - \tau$ for $i \in F$
\If{$u < h$} \Comment{reflect}
    \State $v_j \gets -v_j$ with $j \in A$ drawn w.p. $\propto \lambda_j(\tau)$
\ElsIf{$\beta_i$ reached zero} \Comment{freeze}
    \State move $i$ from $A$ to $F$, \ $v_i \gets 0$, \ $s_i \sim \mathrm{Exp}(\kappa_i)$
\ElsIf{$s_i=0$} \Comment{thaw}
    \State move $i$ from $F$ to $A$, restore $v_i$
\EndIf
\end{algorithmic}
\end{algorithm}

We build on the tangent bound of \citet{andral2024}, but apply it to a whole window rather than a grid of segments. On an interval $[a, b]$, the rate and its time derivative are evaluated at both ends, and
\begin{equation}
\tilde{\lambda} = \max\big(\lambda(a), \lambda(b), m\big),
\end{equation}
where $m$ is the height at which the tangents to $\lambda$ at $a$ and $b$ intersect. A rate that is concave on $[a,b]$ lies below both tangents, and one that is convex lies below the line between $\lambda(a)$ and $\lambda(b)$, so $\tilde\lambda$ bounds the rate whenever it does not switch between concave and convex within $[a,b]$. \citet{andral2024} further divide the window into $K$ segments and bound each one in this way, which gives a tight piecewise constant bound whose $K+1$ evaluations can be computed in parallel. In a BNN, however, every evaluation of the rate and its time derivative costs a full gradient of the log posterior and a Hessian-vector product, and all $K+1$ evaluations are paid before it is known where in the window the event falls. We therefore depart from their scheme and use the tangent bound with a single segment, $K = 1$, letting the window length $t_{\max}$, adapted as in \citet{andral2024}, take the role of the segment width. Building a window then costs two evaluations, and only one when the previous window ended without an event, since its right end is the left end of the next window. This reduces the cost per event by a factor of 1.6 to 2.5 on one of our UCI models (\Cref{sec:supp_grid_vs_singlesegment}). As in \citet{andral2024}, the ZigZag rate is bounded coordinate-wise and summed, while the Boomerang bounds its scalar rate directly.

For a BNN, the rate may switch between concave and convex within a window, so our samplers are exact only when the bound holds at all times in the window. At every proposed time $\tau$, a rate exceeding the bound is detected as a violation, counted and reported, and the window is rebuilt (\Cref{sec:supp_bounds}). The rate is only evaluated at proposed times, so a violation elsewhere in the window goes unnoticed.

\paragraph{Implementation}
\label{sec:implementation}
The only model-specific inputs to our samplers are a standard \texttt{nn.Module} in PyTorch \citep{paszke2019pytorch}, which defines the likelihood, and a prior on its parameters, which together give the energy in \eqref{eq:energy_func}. Any architecture that can be written in PyTorch can therefore be used without changes to the sampler, with support for GPU acceleration. As in \citet{corbella2022} and \citet{andral2024}, the user does not need to derive bounds on the event rates, which is usually the main obstacle to applying PDMPs to new models. Sparsity is controlled by the prior inclusion probabilities $w_i$ alone, from which the sticking rates $\kappa_i$ follow in closed form. Every violation detected at a proposal is counted and reported, so the user can check the output. All code is available at \url{https://github.com/ltronneb/sticky_bnns}.

\section{Experiments}
We now evaluate the PDMP samplers on posterior inference for BNNs. The dense, non-sticky PDMPs target the usual posterior, and here HMC sampling using the No-U-Turn-Sampler (NUTS, \citet{hoffman2014no}) will serve as our baseline. The sticky samplers target the posterior based on the spike-and-slab prior, exploring subspaces that HMC cannot reach. For this model there is no exact sampler to compare with. The stochastic gradient sticky ZigZag of \citet{fearnhead2024stochasticgradientpiecewisedeterministic} discretises time and needs a tuned step size, and we leave a comparison with it to future work. We instead compare with the LBBNN model of \citet{hubin_sparse_bnn}, which approximates the spike-and-slab posterior based on mean field variational inference.

All samplers start at the same reference MAP estimate, as do the variational means of the LBBNN, except NUTS on the one-dimensional data, which uses its default random initialization. The dense models have fan-in scaled Gaussian priors, and the sticky models a spike-and-slab prior with the same Gaussian as slab. Details on the priors, the MAP and the reference measure are given in \Cref{sec:supp_implementation}, and the settings of every experiment in \Cref{tab:supp_hyperparameter_setting}. Each PDMP run is given a fixed budget of either $P$ skeleton points or $G$ gradient evaluations, specified per experiment, where $G$ counts gradients only, not the Hessian-vector products that the bounds also need (\Cref{sec:bounds}). Posterior samples are obtained by evaluating the trajectory at $S=4000$ times drawn uniformly after discarding the first $20\%$ of its time as burn-in. NUTS runs four chains of $1000$ warmup and $1000$ retained draws, giving $S=4000$, and we draw $S=4000$ samples from the fitted LBBNN (\Cref{sec:supp_lbnnn}). For regression, we form the posterior predictive
\begin{equation}
p(y| x) = \frac{1}{S}\sum_{s=1}^S
\mathcal N\!\big(y;\, f(x;\boldsymbol\beta^s),\, \sigma_s^2\big) \ ,
\end{equation}
and report the RMSE of its mean, the NLL and the CRPS on held-out test data (details in \Cref{sec:supp_metrics}).

We first consider a one-dimensional regression problem (\Cref{sec:one_dim}), before turning to the Boston Housing dataset, our main benchmark, where we evaluate predictive performance (\Cref{sec:benchmarks}) and convergence (\Cref{sec:convergence}). Finally, we scale to image classification on MNIST (\Cref{sec:images}). \Cref{sec:supp_additional_experiments} contains additional experiments.

\subsection{A one-dimensional non-linear example}
\label{sec:one_dim}
We fit the PDMP samplers, NUTS and the LBBNN, each in a single run, to a simulated data set using a BNN with one hidden layer of 100 units. The PDMPs use $G=2\times 10^5$ gradient evaluations and the sticky samplers and LBBNN apply a prior inclusion weight $w=0.1$. The responses are
\begin{equation}
\begin{aligned}
\label{eq:toy_dgp}
    y_i &\sim \mathcal N(f(x_i), 0.1^2) \\
    f(x) &= 0.3\tanh(x) + 1.5\exp\{-(x-0.5)^2/0.05\} \ ,
\end{aligned}
\end{equation}
at $40$ inputs drawn uniformly on $[-3,3]$ and $20$ more on $[0,1]$ that oversample the sharp bump of $f$. We call this dataset \emph{sharp}. The metrics in \Cref{tab:toys} are computed on evenly spaced test points in $[-3, 3]$, the range of the training inputs.

\begin{figure}[t]
\centerline{\includegraphics[width=0.7\textwidth]{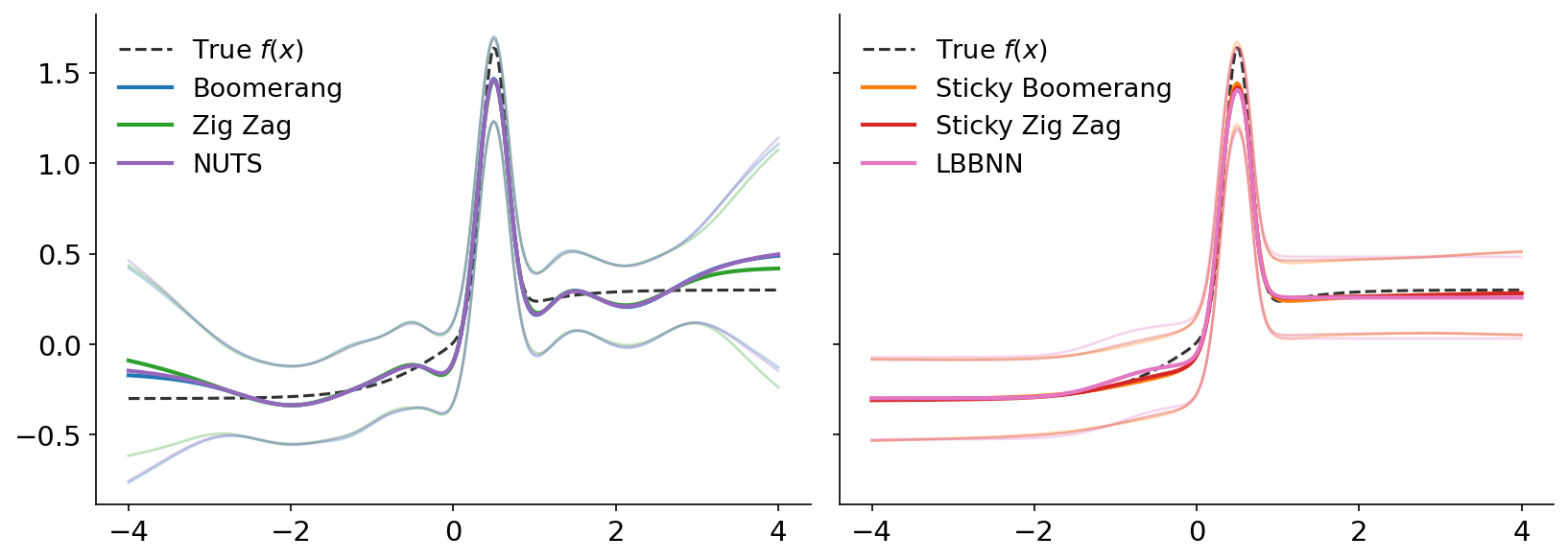}}
\caption{Posterior predictive fits on the \emph{sharp} dataset. The dashed black line is the true function $f$, solid lines are posterior predictive means, and thin lines lie two predictive standard deviations away from the mean.}
\label{fig:toy_bnn_sharp}
\end{figure}

\begin{table}
\centering
\begin{tabular}{lcccc}
\toprule
Sampler & Sparsity & RMSE & NLL & CRPS \\
\midrule
Boomerang        & -- & 0.115 & -0.740 & 0.066 \\
ZigZag           & -- & 0.116 & -0.739 & 0.067 \\
NUTS             & -- & 0.115 & -0.742 & 0.066 \\
Sticky Boomerang & 85\% & \textbf{0.109} & \textbf{-0.801} & \textbf{0.062} \\
Sticky ZigZag    & 87\% & \textbf{0.109} & -0.799 & \textbf{0.062} \\
LBBNN            & 95\% & 0.111 & -0.778 & 0.063 \\
\bottomrule
\end{tabular}
\caption{
Predictive performance on the \emph{sharp} dataset, on evenly spaced test points in $[-3,3]$, for a single run of each method. Sparsity is the percentage of weights that are exactly zero, averaged over draws. Lower is better for RMSE, NLL and CRPS, and the best value is in bold.
}
\label{tab:toys}
\end{table}
As seen in \Cref{fig:toy_bnn_sharp}, all samplers capture the bump, with the peak slightly underestimated. The predictive means of the non-sticky PDMPs and NUTS are visually indistinguishable within the range of the data, which is expected as they target the same posterior. Away from the bump, the dense posteriors follow spurious oscillations on the flat parts of $f$ and widen quickly outside the range of the data. The sticky samplers instead recover the flat, saturating parts of $f$ closely. This is reflected in \Cref{tab:toys}, where the sticky samplers reach a test RMSE of $0.109$, close to the noise level of $0.1$, against $0.115$ to $0.116$ for the dense samplers, and also improve NLL and CRPS. The two sticky samplers agree closely with each other as they target the same posterior, and resemble the LBBNN, which achieves greater sparsity. On the other one-dimensional datasets in \Cref{sec:supp_extended_toys}, the LBBNN performs on par with the sticky samplers on the gap data but worse on the Hernandez and multiscale data, most clearly on multiscale.

\subsection{Benchmark datasets}
\label{sec:benchmarks}
We benchmark on the Boston Housing dataset \citep{harrison1978boston} and three UCI regression datasets (concrete, energy and yacht, see \citet{uci_dataset_name}), using five independent train/test splits per dataset. We consider two BNN architectures, namely a small network with one hidden layer of 50 units and a medium network with three hidden layers of 50 units each. The PDMPs are given a budget of $G=10^6$ gradient evaluations, and the sticky samplers are run with prior inclusion weights $w=0.3$ and $w=0.1$. The LBBNN is given the same spike-and-slab prior with $w=0.3$.

The metrics in \Cref{tab:boston} and in \Cref{sec:supp_extended_UCI} are evaluated on the held-out test set of a 90/10 split, and we additionally report the coverage of the central 90\% predictive interval (\Cref{sec:supp_metrics}).
\begin{table}
\centering
\begin{tabular}{rllccccc}
\toprule
Network & Sampler & $w$ & Sparsity & RMSE & NLL & CRPS & $C_{0.9}$ \\
\midrule
$d=752$
& Boomerang        & --   & --   & 2.88 $\pm$ .19 & \textbf{2.48 $\pm$ .08} & 1.58 $\pm$ .10 & \textbf{90\%} \\
& ZigZag           & --   & --   & 2.88 $\pm$ .22 & 2.49 $\pm$ .08 & 1.58 $\pm$ .12 & 91\% \\
& NUTS             & --   & --   & \textbf{2.86 $\pm$ .22} & \textbf{2.48 $\pm$ .08} & \textbf{1.57 $\pm$ .12} & 91\% \\
& Sticky Boomerang & 0.3  & 64\% & 2.98 $\pm$ .18 & 2.51 $\pm$ .06 & 1.62 $\pm$ .10 & 89\% \\
& Sticky ZigZag    & 0.3  & 67\% & 2.97 $\pm$ .23 & 2.51 $\pm$ .08 & 1.63 $\pm$ .12 & 92\% \\
& Sticky Boomerang & 0.1  & 85\% & 3.07 $\pm$ .19 & 2.54 $\pm$ .06 & 1.67 $\pm$ .10 & \textbf{90\%} \\
& Sticky ZigZag    & 0.1  & 86\% & 3.07 $\pm$ .25 & 2.55 $\pm$ .08 & 1.68 $\pm$ .13 & 92\% \\
& LBBNN            & 0.3  & 86\% & 3.27 $\pm$ .24 & 2.61 $\pm$ .07 & 1.79 $\pm$ .13 & 92\% \\
\midrule
$d=5852$
& Boomerang        & --   & --   & 2.75 $\pm$ .13 & 2.55 $\pm$ .08 & \textbf{1.48 $\pm$ .06} & 82\% \\
& ZigZag           & --   & --   & 2.78 $\pm$ .19 & 2.45 $\pm$ .10 & 1.50 $\pm$ .10 & 88\% \\
& NUTS             & --   & --   & \textbf{2.73 $\pm$ .20} & \textbf{2.44 $\pm$ .09} & \textbf{1.48 $\pm$ .11} & \textbf{89\%} \\
& Sticky Boomerang & 0.3  & 67\% & 2.87 $\pm$ .20 & 2.69 $\pm$ .09 & 1.56 $\pm$ .09 & 81\% \\
& Sticky ZigZag    & 0.3  & 69\% & 2.90 $\pm$ .24 & 2.49 $\pm$ .10 & 1.56 $\pm$ .12 & 88\% \\
& Sticky Boomerang & 0.1  & 87\% & 2.89 $\pm$ .20 & 2.70 $\pm$ .10 & 1.58 $\pm$ .10 & 81\% \\
& Sticky ZigZag    & 0.1  & 89\% & 2.99 $\pm$ .24 & 2.52 $\pm$ .09 & 1.63 $\pm$ .12 & 87\% \\
& LBBNN            & 0.3  & 88\% & 3.62 $\pm$ .28 & 2.72 $\pm$ .07 & 1.99 $\pm$ .15 & \textbf{89\%} \\
\bottomrule
\end{tabular}
\caption{
Predictive performance on the Boston Housing test splits, given as mean $\pm$ standard error (SEM) over 5 independent train/test splits. $d$ is the number of parameters of the small and medium networks. Sparsity is the percentage of weights sampled exactly zero, averaged over draws. $C_{0.9}$ is the share of test responses inside the central 90\% predictive interval, with a  SEM of 2 to 3 percentage points (omitted for space).
}
\label{tab:boston}
\end{table}
In \Cref{tab:boston}, the non-sticky samplers and NUTS have a slight edge over the sticky samplers. The difference in RMSE, NLL and CRPS between the dense and the sticky samplers is small relative to the between-split variation for the small network. It is somewhat larger for the medium network, most clearly in the NLL of the sticky Boomerang. Coverage is close to nominal for all samplers on the small network, while on the medium network all Boomerang samplers produce too narrow predictive intervals, in line with their higher NLL. The sticky samplers sample about two-thirds of the network weights as exactly zero at $w=0.3$, and about $85$ to $89\%$ at $w=0.1$, so the cost in predictive accuracy buys a substantial reduction in the effective dimensionality of the model. The LBBNN is behind all samplers in RMSE and CRPS on both networks, even at matched sparsity ($w=0.3$ for the LBBNN against $w=0.1$ for the sticky samplers). The effect of $w$ on sparsity is studied further in \Cref{sec:supp_sparsity_ablation}, and results for the remaining UCI datasets are given in \Cref{sec:supp_extended_UCI}. They show essentially the same pattern as for Boston, with the sticky samplers comparable to the dense PDMPs and NUTS, the LBBNN performing worse on every dataset, and the Boomerang samplers on the medium network giving too narrow intervals on concrete. Lastly, no bound violations were detected in any of the one-dimensional or benchmark runs.

\subsection{Convergence analysis}
\label{sec:convergence}
\begin{figure}[t]
\centerline{\includegraphics[width=0.7\textwidth]{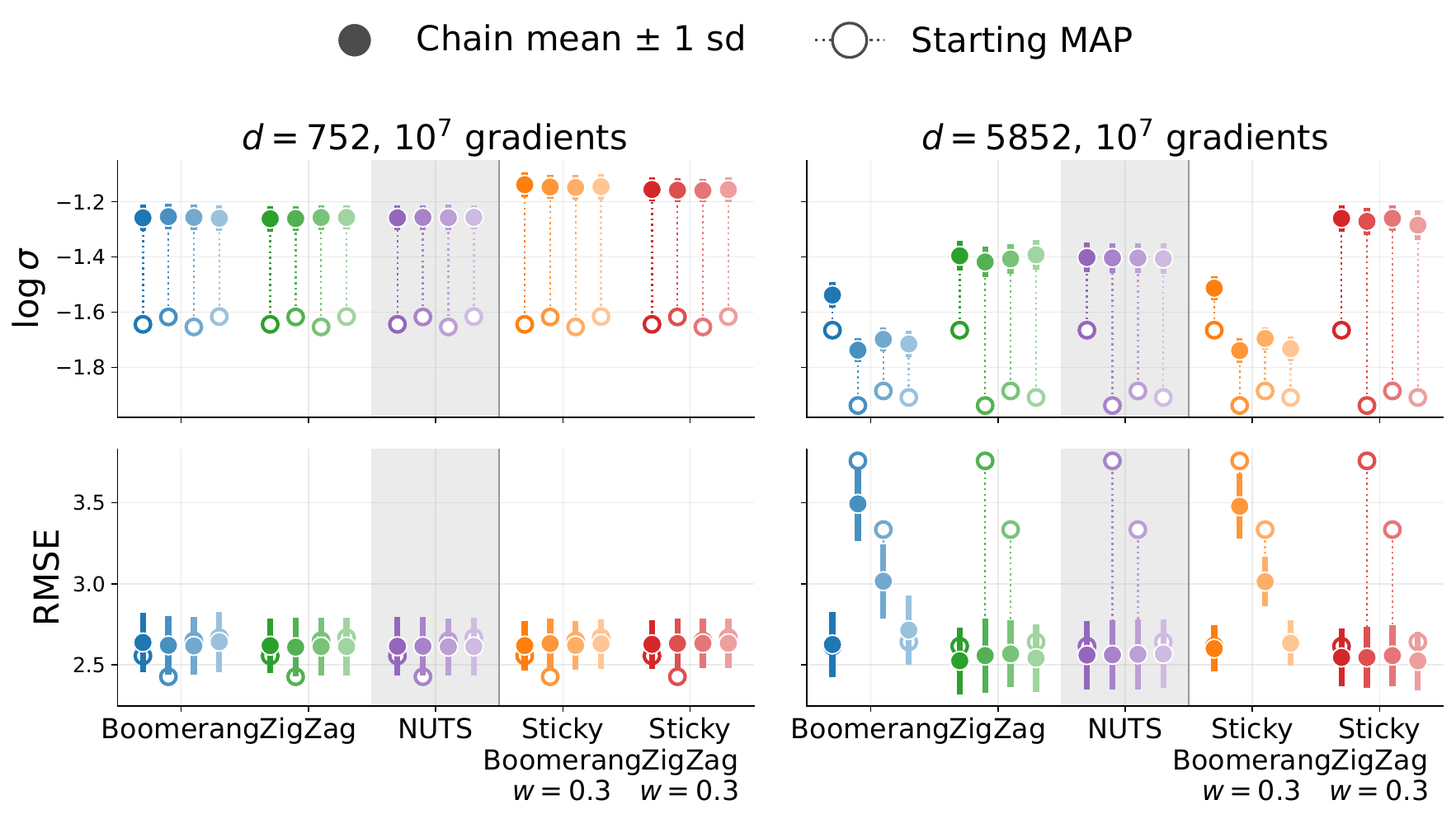}}
\caption{Chain means of $\log\sigma$ (top) and test RMSE (bottom) on the first Boston split. Filled dots show each chain's mean with one posterior standard deviation, and hollow dots the MAP it started from.}
\label{fig:convergence}
\end{figure}

To assess convergence, we run four chains of every sampler in \Cref{tab:boston} on the first of the five train/test splits. We fit four MAP estimates from different random initialisations, which end in different local modes of the multimodal posterior, and start one chain from each, with the Laplace covariance at that MAP. The PDMPs are given a budget of $G=10^7$ gradient evaluations per chain, and NUTS starts from the same four MAPs. We compare the chains on the noise level $\log \sigma$ and test RMSE, and summarise their agreement with the rank-normalised $\hat R$ of \citet{Vehtari_rank_norm_rhat}. On the small network, every sampler converges (\Cref{fig:convergence}), with $\hat R \le 1.01$ and ESS larger than 400 for both statistics except for the sticky Boomerang at $w=0.1$ as seen in \Cref{tab:supp_rhat}. On the medium network, the ZigZag chains agree ($\hat R \leq 1.04$) and reach the same $\log \sigma$ and test RMSE regardless of MAP. However, with a bulk ESS of 70 to 170, they mix slowly and we do not claim full convergence. The Boomerang samplers on the medium network do not converge and seem stuck close to whatever MAP they start off in. We attribute this to the Gaussian reference of the Boomerang (\Cref{sec:supp_reference_measure}). A MAP that overfits the training data will underestimate the noise level $\sigma$, and since the Laplace precision of the weights scales roughly with $1/\sigma^2$, the reference becomes narrow. The Boomerang then moves on small orbits around the MAP, so the chain stays close to where it started. Consistent with this, all four MAPs underestimate the noise level (\Cref{fig:convergence}), and while the ZigZag and NUTS chains all rise to the same $\log\sigma$, the Boomerang chains only move part of the way and settle below it. NUTS converges on both networks as expected, and is more effective per gradient evaluation.

The convergence analysis matters for \Cref{tab:boston} in two ways. First, for the small network, and the ZigZag samplers on the medium network, the test RMSE is already close to its final value at the budget of $G=10^6$ used in the table, while $\log\sigma$ mixes more slowly (\Cref{fig:traces}), consistent with the finding of \citet{papamarkou2022challenges} that non-converged chains can still predict accurately. Furthermore, the table averages over five splits, and the variation between splits is several times larger than the variation between chains. The Boomerang samplers on the medium network are the exception, as their chains differ about as much as the splits do. More diagnostics and details can be found in \Cref{sec:supp_convergence}.

\subsection{Image data}
\label{sec:images}
\begin{figure}[t]
\centering
\begin{minipage}[c]{0.37\textwidth}
    \centering
    \includegraphics[width=\linewidth]{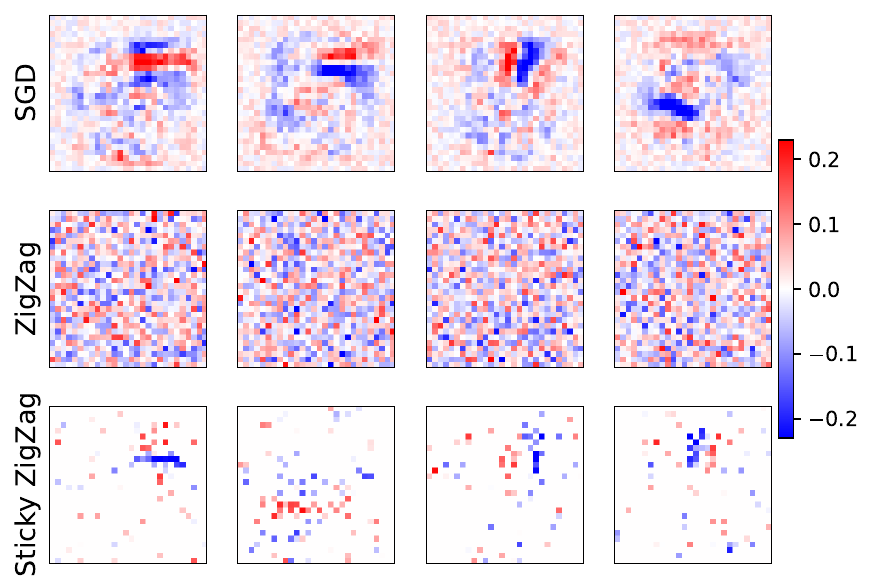}\\
    \small (a)
\end{minipage}
\hfill
\begin{minipage}[c]{0.62\textwidth}
    \centering
    \includegraphics[width=\linewidth]{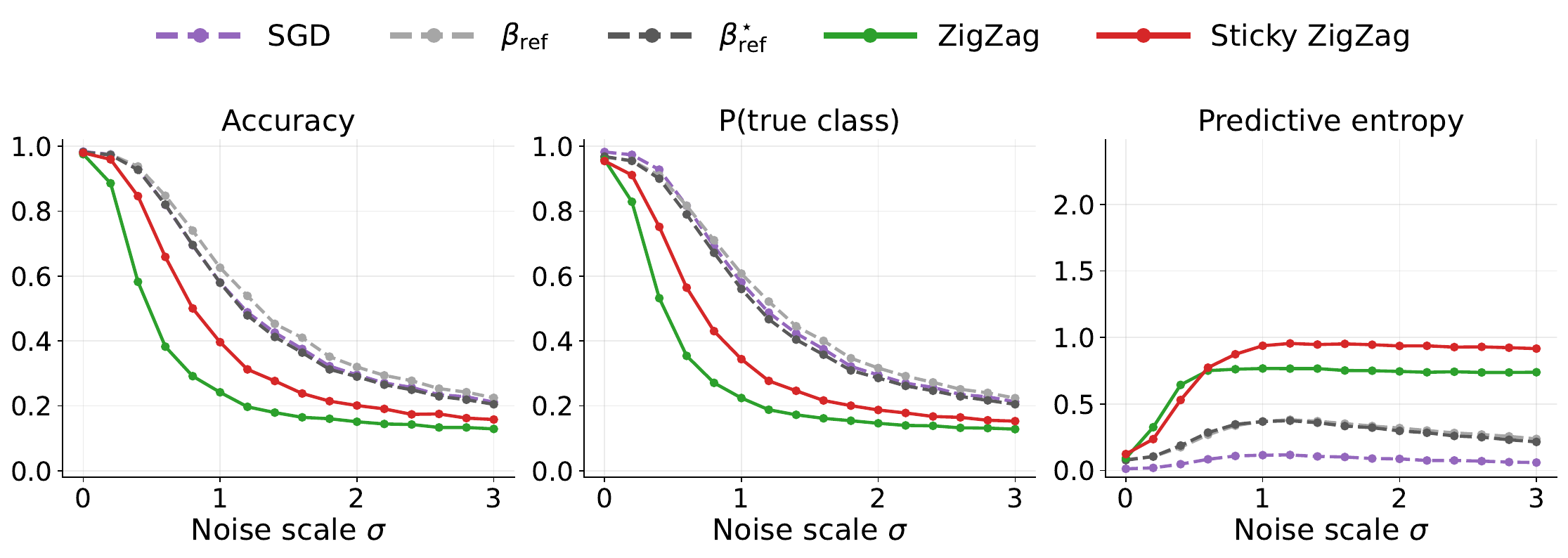}
    \small (b)
\end{minipage}
\caption{Analyses following \citet{izmailov2021dangers} (a) and \citet{izmailov_hmc_bnn} (b), adapted to our samplers for the MNIST FFN. (a) First-layer weights of four hidden units. (b) Robustness to Gaussian noise of standard deviation $\sigma$ on the $[0,1]$ pixel scale.}
\label{fig:MNIST_noise}
\end{figure}

To demonstrate scalability, we apply the dense and sticky ZigZag to MNIST \citep{LeCun_LeNet5}, with a feedforward network (FFN) with two hidden layers of 256 ReLU units and a LeNet-5 with $\tanh$ activations. Each sampler is run once for $P=10^6$ skeleton points with minibatch gradients of size 1024, which makes the samplers approximate, and we briefly discuss the effects of minibatching in \Cref{sec:supp_minibatch}. The sticky ZigZag starts at a pruned MAP, $\boldsymbol{\beta}_{\mathrm{ref}}^{\star}$, in which weights that can be set to zero without degrading the validation accuracy start frozen, and the dense ZigZag at the unpruned MAP $\boldsymbol{\beta}_{\mathrm{ref}}$. We compare with an SGD baseline, which is described in \Cref{sec:supp_baselines}. As we do not expect the samplers to converge on these networks, we focus on the ZigZag, which converged most reliably on the UCI data (\Cref{sec:convergence}). Furthermore, with ReLU activations the rate can jump discontinuously, so the tangent bound carries no guarantee for the FFN. Neither ZigZag sampler reports any bound violations, while the sticky Boomerang, reported in \Cref{sec:supp_mnist}, records 71 and 79 bound violations for the FFN and LeNet-5.

On the test images, all samplers stay within 0.8 percentage points of the accuracy of SGD, while the sticky ZigZag sets 92 to 98\% of the weights exactly to zero (\Cref{tab:MNIST}). \citet{izmailov_hmc_bnn} find that BNNs sampled with HMC are less robust than SGD to Gaussian noise on MNIST, and \citet{izmailov2021dangers} attribute this to pixels that are zero in all training images, as the posterior of the weights connected to them equals the prior. We adapt both analyses to our samplers in \Cref{fig:MNIST_noise}. In our draws, the weights of these pixels are indeed at their prior, but under the spike-and-slab prior only $5\%$ of them are nonzero, against all of them for the dense ZigZag. More broadly, the first-layer weights of the dense ZigZag look like noise, while the nonzero weights of the sticky ZigZag form stroke-like patterns similar to those of SGD (\Cref{fig:MNIST_noise}a). Under unclipped Gaussian noise, the sticky ZigZag is clearly more robust than the dense ZigZag, which these weight patterns may partly explain, although it remains less robust than SGD and both MAPs (\Cref{fig:MNIST_noise}b). On the LeNet-5, whose filters are shared across pixel positions, the two samplers behave alike (\Cref{fig:supp_lenet}).

\section{Discussion}
In this work we have shown that sticky PDMPs can sample the posterior of BNNs under a spike-and-slab prior. The samplers set a large share of the weights exactly to zero without thresholding or other post-processing, at a modest cost in predictive performance compared with dense samplers. The samplers predict more accurately than a sparse variational approximation under the same prior on all our benchmark datasets.

To make PDMPs more efficient, we bound the rate with a cheap single tangent segment over an adaptive window, valid whenever the rate does not switch between concave and convex within it. The samplers only need a differentiable log density, specified as a PyTorch module, and can run on GPUs. On the UCI benchmarks we study convergence across chains started in different MAPs, and on MNIST we show that sticky PDMPs can make a sparse feedforward BNN more robust to Gaussian input noise than a dense one.

\paragraph{Limitations}
In the convergence analysis on the first Boston split (\Cref{sec:convergence}), the PDMPs mix more slowly than NUTS per gradient evaluation, for the dense ZigZag by a factor of about 20 on the small and 50 on the medium network in effective sample size of $\log\sigma$. The comparison is approximate, as the effective sample sizes come from draws rather than the continuous trajectory and the gradient count leaves out the Hessian-vector products, but neither explains a gap of this size. On the medium network the ZigZag chains agree but mix slowly, and the Boomerang samplers do not converge, as their reference becomes too narrow when the MAP overfits. The test RMSE stabilises before $\log\sigma$ mixes, so predictive results are reached before full convergence. Exactness also rests on bounds that cannot be verified for a BNN, and the MNIST results are single minibatch runs that we do not expect to have converged. Finally, all chains start at a MAP and explore the posterior locally, so a single chain cannot be expected to visit several modes.

\paragraph{Further work}
Several directions could improve PDMP samplers. We believe the largest and simplest gain lies in the velocities of the ZigZag. All coordinates currently move at unit speed, although their posterior scales differ widely. Speeds proportional to the standard deviations of the Laplace approximation at the MAP, or learned during sampling by the adaptive preconditioning of \citet{bertazzi2022}, could account for this. Bounds that remain valid with subsampled gradients, for instance through statistically estimated bounds as in \citet{corbella2022}, could remove the bias of minibatching and significantly lower the cost per event. For the Boomerang, a reference that adapts during sampling and accounts for the frozen weights could reduce its dependence on the MAP. On the modelling side, a hyperprior on $w$, structured priors that freeze whole units, or input-skip connections \citep{hoyheim2026explainable} could make the networks even sparser. Finally, scaling to larger architectures and developing Python software of the kind NumPyro provides for HMC would make PDMPs more practical for Bayesian deep learning.

 \subsection*{AI use statement}

In this work, we used generative AI tools for feedback on research methodology, design and feedback on experiments, implementing methods, formulating mathematical claims, translation, cleaning and reformatting of datasets and in interpreting results. We have not used generative AI tools for generating synthetic datasets, designing research methodology, developing theoretical models and conceptual frameworks, providing critical ingredients for proving mathematical claims, assisting in writing proofs, proposing or refining hypotheses. Support in qualitative and thematic data analysis is not applicable to this work. The invariance argument in \Cref{sec:supp_sticky_boomerang} was discussed with an AI tool and checked by the authors.

Additionally, we used generative AI tools for creating or modifying scientific figures, images and plots, creating and editing software code, drafting parts of a paper, summarizing and analyzing existing literature, brainstorming, sourcing and searching for information, editing text to improve readability, identifying relevant literature and suggesting a structure for a research paper.

We have reviewed all AI-assisted work. The authors have checked and discussed all research methodology present in the paper, and are confident it is sound. The implemented methods have been verified with existing literature rigorously and LLM generated code has been subject to numerical verification to known ground truths and code produced by the authors. All literature-related output from LLMs was checked against the original sources. Additionally, we did not rely on any LLM-provided information that could not be traced to a source we were able to verify. All text drafts and edits have been verified to reflect what the authors wish to express.

We take responsibility for the final content of this work, including text, claims, or artifacts produced with the aid of generative AI.

\section*{Acknowledgements}

The authors wish to thank Aliaksandr Hubin and Alice Corbella for the fruitful discussions.
AA and GOS were supported by the Research Council of Norway through its Centre of Excellence Integreat - The Norwegian Centre for knowledge--driven machine learning, project number 332645.
FP was supported by the UKRI grant EP/Y014650/1 as part of the ERC Synergy project OCEAN.
LR was supported by the European Union's Horizon Europe research and innovation programme under the Marie Sk\l odowska-Curie grant agreement No.\ 101126636.
\bibliographystyle{apalike}
\bibliography{bib}

\clearpage
\appendix

\section{Obtaining samples from the continuous trajectory}\label{sec:supp_samples}
Given a sampled trajectory $\{\boldsymbol{\beta}_t\}_0^T$, we discard the first $20\%$ of the trajectory time as burn-in and denote by $T_0$ the start of the remaining part. The optimal estimator for $\mu_f\equiv E[f(\boldsymbol{\beta})]$ is then
\begin{align*}
\hat\mu_f=\frac{1}{T-T_0}\int_{T_0}^Tf(\boldsymbol{\beta}_t)dt
=\frac{1}{T-T_0}\sum_{j=1}^k\int_{\tau^{j-1}}^{\tau^j}f(\boldsymbol{\beta}_t)dt,
\end{align*}
where $T_0=\tau^0<\tau^1<\cdots<\tau^k=T$ are the skeleton points after the burn-in and where the $k$ integrals to the right are over deterministic paths. When these integrals are difficult to compute, as for the predictive quantities of a BNN, we instead select sample points $t_1,\dots,t_S$ and use~\eqref{eq:pred} with $\boldsymbol{\beta}^s=\boldsymbol{\beta}_{t_s}$. One option is an equally spaced grid on $[T_0,T]$, as in \citet{bierkens2019bigdata}, which gives a Riemann-sum approximation of $\hat\mu_f$. In all our experiments we instead draw $t_1,\dots,t_S$ independently and uniformly on $[T_0,T]$, and sort them into time order. Then $E[f(\boldsymbol{\beta}_{t_s})\mid\{\boldsymbol{\beta}_t\}]=\hat\mu_f$ for every $s$, so the sample average is an unbiased estimate of $\hat\mu_f$, and, unlike a grid, the scheme carries over to the case below where $T$ is not known in advance. Note that we can rewrite $\hat\mu_f$ to
\begin{align*}
\hat\mu_f=\sum_{j=1}^k\frac{T_j}{T-T_0}\frac{1}{T_j}\int_{\tau^{j-1}}^{\tau^j}f(\boldsymbol{\beta}_t)dt,
\end{align*}
where $T_j=\tau^j-\tau^{j-1}$. A time drawn uniformly on $[T_0,T]$ is therefore equivalently obtained by first choosing an interval $[\tau^{j-1},\tau^j)$ with probability $T_j/(T-T_0)$ and then drawing a time uniformly within it.

For the one-dimensional models and the small UCI networks, the entire skeleton is kept in memory, so $T$ is known when sampling and the times are drawn directly. For the larger models, keeping the skeleton in memory can pose issues. Their skeletons are generated in time-segments $[s_{m-1}, s_{m}], m=1, \dots M,$ with $s_0=0$ and $s_M=T$, where each segment is discarded from memory before the next is generated, so $T_0$ and $T$ are not known while the segments are processed. We can however generate uniformly distributed time samples by relying on reservoir sampling \citep{chao_general_1982}. Consider retaining $R$ slots, each holding a time and the corresponding state of the PDMP. For the first segment, fill the slots with times drawn uniformly on $[0, s_1]$ and the corresponding states. Before generating the next segment, the skeleton of the current one is discarded, except its last point, from which the next segment continues. When segment $[s_{m-1}, s_m]$ has been generated, each of the $R$ slots is independently replaced, with probability $(s_m-s_{m-1})/s_m$, by a time drawn uniformly on $[s_{m-1}, s_m]$ and the corresponding state. If a slot held a time uniform on $[0,s_{m-1}]$ before this update, it holds a time uniform on $[0,s_m]$ after it, so by induction the slots hold independent times drawn uniformly on $[0,T]$ once all $M$ segments are processed. We then discard the slots with times before $T_0=0.2\,T$. The remaining slots hold independent times drawn uniformly on $[T_0,T]$, and we keep $S$ of them at random. The resulting draws have the same distribution as if $t_1,\dots,t_S$ had been drawn uniformly on $[T_0,T]$ directly, but only one segment of the skeleton is held in memory at a time.

\section{Implementation details}
\label{sec:supp_implementation}
For an input $\mathbf{x} \in \mathbb{R}^p$, the network output is defined
recursively by the forward pass
\begin{equation}
\begin{aligned}
    f_1(\mathbf{x}; \boldsymbol{\beta}) &= W_1\mathbf{x} + b_1 \ , \\
    f_{\ell}(\mathbf{x}; \boldsymbol{\beta}) &= W_{\ell}\,\varphi\big(f_{\ell-1}(\mathbf{x}; \boldsymbol{\beta})\big) + b_{\ell} \ , \qquad \ell = 2, \dots, L \ ,
\end{aligned}
\end{equation}
where $\varphi$ is a non-linear activation applied elementwise, $W_\ell \in \mathbb{R}^{h_\ell \times h_{\ell-1}}$ and $b_\ell \in \mathbb{R}^{h_\ell}$ are the weight matrix and bias vector of layer $\ell$, $h_0 = p$, and $\boldsymbol{\beta} = \{W_\ell, b_\ell\}_{\ell=1}^L$. We use $\varphi = \tanh$ for all models except the MNIST feedforward network, which uses ReLU. The network output $f_L(\mathbf{x}; \boldsymbol{\beta}) \in \mathbb{R}^{h_L}$ parametrizes a task-dependent likelihood $p(\mathbf{y} \mid f_L(\mathbf{x}; \boldsymbol{\beta}))$.

\subsection{Model specifications}
\label{sec:supp_model_specifications}
Classification tasks on MNIST are assigned a categorical likelihood. Regression tasks are assigned a Gaussian likelihood, where the network output serves as the mean and an additional noise scale $\sigma$ is introduced. For the one-dimensional datasets, the noise scale is fixed at the true value, and for all other regressions it is learned. For non-sticky PDMPs and NUTS, the prior over all network weights is a fan-in scaled Gaussian prior $\mathcal{N}(0, \sigma_{\ell}^2)$ with $\sigma_\ell^2 = \sigma_w^2/F_{\ell}$, where $F_{\ell}$ is the number of inputs to layer $\ell$ and $\sigma_w$ is a prior scale. For sticky PDMPs we apply a spike and slab prior on the network weights with a prior inclusion weight $w$ and a Gaussian slab
\begin{equation}
    p(\beta) = w\mathcal{N}(\beta; 0, \sigma_{\ell}^2) + (1-w)\delta_0(\beta) \ .
\end{equation}
This spike and slab prior is parametrized via the sticking parameter $\kappa$ determined by $w$ and the slab, which is described more in detail in \Cref{sec:supp_sparsity_ablation}.
For all samplers, the bias terms are not freezable and given a Gaussian prior
\begin{equation}
    p(b) = \mathcal{N}(0, \sigma_{b}^2) \ ,
\end{equation}
without a fan-in scale. Collecting the weights and biases, the prior covariance matrix $\Sigma_0$ is diagonal, with $(\Sigma_0)_{ii} = \sigma_w^2/F_{\ell}$ for weights and $(\Sigma_0)_{ii} = \sigma_b^2$ for biases. In the implementation, the parameters of the \texttt{nn.Module} are stored as named tensors, one per layer, so these priors are assigned layerwise before the parameters are flattened into the single vector $\boldsymbol{\beta}$. For regression, the noise scale $\sigma$ of the likelihood is excluded from freezing, since it must be positive. It is given a half-normal prior with scale $s$ chosen per dataset. The values of $\sigma_w$, $\sigma_b$ and $s$ for each experiment are given in \Cref{tab:supp_hyperparameter_setting}.

\subsection{MAP and initialisation}
\label{sec:supp_map}
For each dataset and split, we compute a MAP estimate of a neural network with the same architecture as the relevant BNN. The MAP is found by minimising, up to a normalizing constant,
\begin{equation}
\mathcal L(\boldsymbol{\beta}) = -\log p(\mathcal D \mid \boldsymbol{\beta}) - \log p(\boldsymbol{\beta})
= -\sum_{n=1}^{N} \log p(y_n \mid x_n, \boldsymbol{\beta}) + \tfrac12 \boldsymbol{\beta}^{\top} \Sigma_0^{-1}\boldsymbol{\beta}
\end{equation}
with the Adam optimizer using either a constant learning rate for simpler targets, or a cosine-annealed learning rate for more complex models. For the image models, the MAP is then pruned. Weights with $|\beta_i|$ below $m$ times their prior standard deviation are set to zero, for the largest $m$ on a logarithmic grid in $[10^{-3},1]$ for which the accuracy on $2000$ test images, that are left out of all reported metrics, stays within a tolerance of the unpruned MAP ($1$ and $3$ percentage points for LeNet-5 and the feedforward network). The remaining weights are then refitted with the pruned ones fixed at zero, and the sticky samplers start from this pruned MAP with the pruned weights frozen.

All samplers are initialized at the MAP, in order to get a fair comparison. The MAP also serves as the centre of the Boomerang reference measure (\Cref{sec:supp_reference_measure}). Starting at the MAP is analogous to the notion of control variates in the PDMP literature \citep{bierkens2019bigdata, bierkens2020Boomerang}. The drawbacks are of course that the samplers will be initialized within a mode of the multimodal loss-landscape. This might hinder full exploration of the state space, but for complex tasks the MAP is also essential for a well behaving sampler.

\section{The Boomerang sampler}\label{sec:supp_Boomerang}
In the main body we also included results from sampling with the Boomerang sampler, which is described in the following. Instead of linear motion with constant velocities, the Boomerang sampler is defined on $\mathbb{R}^d \times \mathbb{R}^d$ by the elliptical flow
\begin{equation}\label{eq:Boomerang}
    \Phi_t(\boldsymbol{\beta}, \boldsymbol{v}) =
    \begin{pmatrix}
        \boldsymbol{\mu} + (\boldsymbol{\beta} - \boldsymbol{\mu})\cos t + \boldsymbol{v}\sin t \\[2pt]
        \boldsymbol{v}\cos t - (\boldsymbol{\beta} - \boldsymbol{\mu})\sin t
    \end{pmatrix} .
\end{equation}
Here $\boldsymbol{\mu}$ is the mean and $\Sigma$ the covariance of a Gaussian reference measure $\mathcal{N}(\boldsymbol{\mu}, \Sigma)$, typically chosen to resemble the posterior and $\Sigma$ defines the velocity distribution $\boldsymbol{v} \sim \mathcal{N}(\mathbf{0}, \Sigma)$. By adopting this reference, the posterior can be written as
\begin{equation}\label{eq:Boomerang_target}
\begin{aligned}
    \pi(\boldsymbol{\beta}) &\propto e^{-E(\boldsymbol{\beta})}\,
    \mathcal{N}(\boldsymbol{\beta}; \boldsymbol{\mu}, \Sigma) \ , \\
    E(\boldsymbol{\beta}) &= U(\boldsymbol{\beta})
    - \tfrac{1}{2}(\boldsymbol{\beta} - \boldsymbol{\mu})^\top \Sigma^{-1}
    (\boldsymbol{\beta} - \boldsymbol{\mu}) \ .
\end{aligned}
\end{equation}
The flow rotates $(\boldsymbol{\beta}-\boldsymbol{\mu}, \boldsymbol{v})$ and thus leaves $\mathcal{N}(\boldsymbol{\mu}, \Sigma) \otimes \mathcal{N}(\mathbf{0}, \Sigma)$ invariant, so the Gaussian part of the target is handled exactly and $E$ only measures the departure of the posterior from the reference. If the posterior is close to Gaussian, $\nabla E$ will be small and events are rare, meaning sampling is very efficient. More details on how we choose $\boldsymbol{\mu}$ and $\Sigma$ can be found in \Cref{sec:supp_reference_measure}. In contrast to the ZigZag, the rate of the Boomerang does not factorise, and takes the general form
\begin{equation}
    \lambda(t) = \langle \boldsymbol{v}(t), \nabla E(\boldsymbol{\beta}(t)) \rangle_+ \ .
\end{equation}
To sample event times, we simply perform thinning on $\lambda(t)$ directly. When an event occurs, the Boomerang transition is a contour reflection defined by
\begin{equation}
\mathbf{R}(\boldsymbol{\beta}) \boldsymbol{v} = \boldsymbol{v} - \frac{2\langle \nabla E(\boldsymbol{\beta}), \boldsymbol{v} \rangle}{\lvert \Sigma^{1/2}\nabla E(\boldsymbol{\beta}) \rvert^2} \Sigma \nabla E(\boldsymbol{\beta}) \ .
\end{equation}
Contour reflections alone do not make the Boomerang sampler ergodic \citep{bierkens2020Boomerang}. Velocities are therefore refreshed by drawing $\boldsymbol{v} \sim \mathcal{N}(\mathbf{0}, \Sigma)$ at event times sampled from a homogeneous Poisson process with rate $\lambda_{\mathrm{ref}}$.

\subsection{Reference measure for BNNs}
\label{sec:supp_reference_measure}
The Boomerang samplers require a Gaussian product measure $\mathcal N(\boldsymbol{\mu}, \Sigma) \otimes \mathcal{N}(\mathbf{0}, \Sigma)$, and so we must find candidates for $\boldsymbol{\mu}$ and $\Sigma$. We set $\boldsymbol{\mu}$ to the MAP of \Cref{sec:supp_map}, and obtain $\Sigma$ from a Laplace approximation at the MAP. For the reference covariance, we use a diagonal approximation that combines the prior precision $\Sigma_0^{-1}$ of \Cref{sec:supp_model_specifications} and the diagonal empirical Fisher information. Concretely, we calculate
\begin{equation}
(\Sigma^{-1})_{ii} = (\Sigma_0^{-1})_{ii} +
\frac{N}{M}\sum_{m=1}^{M}
\left(
\frac{\partial}{\partial\beta_i}\log p\!\left(y_{n_m}\mid x_{n_m},\boldsymbol{\beta}\right)
\right)^{\!2}_{\boldsymbol{\beta}=\boldsymbol{\mu}},
\qquad i = 1,\dots,d \ ,
\end{equation}
for a uniform sample $\{n_1,\dots,n_M\} \subset \{1, \dots, N\}$ without replacement. The factor $N/M$ turns the $M$-point sum into an unbiased estimate of the full-data Fisher sum $\sum_{n=1}^N(\partial_{\beta_i}\log p(y_n\mid x_n,\boldsymbol{\beta}))^2$, so that $\Sigma^{-1}$ is comparable across training-set sizes. Each diagonal entry is floored at $\epsilon = 10^{-8}$.

In practice the Boomerang sampler's mixing is sensitive to the overall scale of the reference measure. We therefore multiply the reference precision $\Sigma^{-1}$ by a scalar $c>0$, set by hand by inspecting how $\sigma$ developed along pilot trajectories (\Cref{tab:supp_hyperparameter_setting}). Values $c < 1$ loosen the reference measure, while $c > 1$ tighten it. For regression tasks, the entry of $\Sigma^{-1}$ corresponding to $\log\sigma$ is set to the curvature of the log noise prior at the MAP, $2\sigma_{\mathrm{ref}}^2/s^2$ for a half-normal prior with scale $s$, instead of the empirical Fisher term, since the likelihood curvature in $\log\sigma$ ignores its coupling to the weights and makes the reference overly concentrated. This entry is not multiplied by $c$.

\subsection{Sticky Boomerang sampler}
\label{sec:supp_sticky_boomerang}
The sticky Boomerang is similar to the sticky ZigZag and we follow \citet{bierkens2022sticky}. When an active coordinate $\beta_i$ hits zero, it freezes with its velocity $v_i$ stored, and thaws after an exponential time with rate $\kappa_i \lvert v_i\rvert$, continuing with the stored velocity. While frozen, it is excluded from the event rate and from reflections. At a refreshment, a frozen coordinate keeps the sign of its stored velocity and redraws its magnitude \citep{bierkens2022sticky}, together with its thawing time. Unlike \citet{bierkens2022sticky}, who centre the reference at zero, we centre it at the MAP with a diagonal covariance. The centre only changes how the active coordinates move, not the sticky dynamics, and in stationarity the rates of freezing and thawing at zero are set by the target density there. We therefore take the invariance argument of \citet[Proposition~B.3]{bierkens2022sticky} to carry over to centring the sticky Boomerang at the MAP.

\section{Experimental settings}
\label{sec:supp_experiment_setting}
\subsection{Data}
\label{sec:supp_data}
The Boston Housing and UCI datasets are split at random into $90\%$ training and $10\%$ test data, with the five splits drawn with seeds 42 to 46. Inputs and responses are standardised with the mean and standard deviation of the training split, and all metrics are reported on the original scale. The one-dimensional datasets are generated as described in \Cref{sec:one_dim} and \Cref{sec:supp_extended_toys}, with test responses drawn from the same model at $300$ evenly spaced points in $[-3, 3]$ ($[-4, 4]$ for Hernandez). The image models use the full MNIST training sets of $60{,}000$ images. Of the $10{,}000$ test images, $2000$ are used to choose the pruning threshold of the MAP (\Cref{sec:supp_map}), and all reported metrics are computed on the remaining $8000$. We recognise the ethical concerns with the Boston Housing dataset, and use it only because it remains a widely used benchmark, formerly part of the UCI repository. For brevity, we sometimes count it among the UCI datasets in the text.

\subsection{Hyperparameters}
\label{sec:supp_hyperparams}
\Cref{tab:supp_hyperparameter_setting} lists the settings of every experiment. The prior scales $\sigma_w$ and $\sigma_b$ were fixed in advance for each experiment. On the UCI data, the prior inclusion weights $w=0.3$ and $w=0.1$ give two levels of sparsity, and \Cref{sec:supp_sparsity_ablation} covers the full range of $w$. The image models use smaller $w$, given their size. The reference scale $c$ was tuned per dataset for the Boomerang. Furthermore, in practice, the ZigZag adds a constant excess switching rate $\gamma_{\mathrm{const}}$ to the rate of each moving coordinate, $\lambda_i = (v_i \partial_i U)^+ + \gamma_{\mathrm{const}}$, which ensures ergodicity \citep{bierkens2019bigdata}. The initial $t_{\max}$ only sets the starting point of the adaptation in \Cref{sec:supp_bounds}, and is not critical. The number of Adam steps for the MAP was chosen so that the training loss had flattened.
\begin{table}
\centering
\resizebox{\columnwidth}{!}{
\begin{tabular}{lccccc}
\toprule
 & One-dimensional & UCI small & UCI medium & MNIST FFN & MNIST LeNet-5 \\
\midrule
Hidden layers & $[100]$ & $[50]$ & $[50, 50, 50]$ & $[256, 256]$ & LeNet-5  \\
Activation & $\tanh$ & $\tanh$ & $\tanh$ & ReLU & $\tanh$ \\
Likelihood & Gaussian, $\sigma$ fixed & Gaussian, $\sigma$ learned & Gaussian, $\sigma$ learned & Categorical & Categorical \\
$\sigma_w$, $\sigma_b$ & 3, 3 & 1, 1 & 1, 1 & 2, 2 & 2, 2 \\
Noise prior scale $s$ & -- & per dataset$^\star$ & per dataset$^\star$ & -- & -- \\
Inclusion weight $w$ & 0.1 & 0.3, 0.1 & 0.3, 0.1 & 0.05 & 0.05  \\
Gradient batch & full & full & full & 1024 & 1024 \\
ZigZag base rate $\gamma_{\mathrm{const}}$ & $10^{-2}$ & $10^{-2}$ & $10^{-2}$ & $10^{-6}$ & $10^{-6}$ \\
Boomerang refresh rate & 1 & 1 & 1 & 1 & 1 \\
Reference scale $c$ & 0.1 & 0.1 & per dataset$^\star$ & 1 & 1 \\
Initial $t_{\max}$, ZigZag & $2\cdot 10^{-4}$ & $2\cdot 10^{-4}$ & $2\cdot 10^{-4}$ & $10^{-5}$ & $10^{-5}$  \\
Initial $t_{\max}$, Boomerang & $3\cdot 10^{-3}$ & $3\cdot 10^{-3}$ & $3\cdot 10^{-3}$ & $10^{-3}$ & $10^{-3}$ \\
Adaptation $\alpha$ & 1.01 & 1.01 & 1.01 & 1.02 & 1.02 \\
$t_{\max}$ shrink after a violation & 2 & 2 & 2 & 1.1 & 1.1 \\
MAP, Adam steps & $10^4$ & $2\cdot 10^4$ & $2\cdot 10^4$ & $10^4$ & $10^4$ \\
MAP, learning rate & $10^{-2}$ & $10^{-2}$ & $10^{-2}$ & $2\cdot 10^{-3}$, cosine & $2\cdot 10^{-3}$, cosine \\
MAP, batch & full & full & full & full & 512 \\
Fisher sample size $M$ & 128 & 128 & 128 & 1024 & 1024  \\
\bottomrule
\end{tabular}
}
\caption{Settings per experiment. $^\star$Per dataset, in the order Boston, Energy, Concrete and Yacht, $s = 0.3$, $0.03$, $0.2$, $0.01$ and $c = 0.75$, $0.5$, $10$, $0.5$.
}
\label{tab:supp_hyperparameter_setting}
\end{table}

\subsection{Performance metrics}
\label{sec:supp_metrics}
We here include more details on how the predictive metrics are computed. The responses are standardised with the training mean and standard deviation $\hat\sigma_y$, and all metrics are reported on the original scale. Write $\mu_n^{(s)} = f_{\boldsymbol{\beta}^s}(x_n)$ and $\sigma^{(s)}$ for the mean and noise level of draw $s$ at test point $n$. The reported RMSE and NLL are calculated as
\begin{equation}
\mathrm{RMSE} = \hat\sigma_y\sqrt{\frac1N\sum_{n=1}^N\big(\bar f(x_n) - y_n\big)^2} \qquad \mathrm{NLL} = -\frac{1}{N}\sum_n \log\left(\frac{1}{S}\sum_s p(y_n\mid x_n, \boldsymbol{\beta}^s)\right)
\end{equation}
with $\bar f(x) = \frac1S\sum_s f_{\boldsymbol{\beta}^s}(x)$. The continuous ranked probability score (CRPS) of a predictive distribution with CDF $F$ at an observation $y$ is
\begin{equation}
\mathrm{CRPS}(F, y) = \int_{-\infty}^{\infty} \big(F(z) - \mathbf 1\{y \le z\}\big)^2 \,\mathrm dz \ ,
\end{equation}
where lower is better. Since each draw $\boldsymbol{\beta}^s$ induces a Gaussian predictive, the posterior predictive is an equally weighted Gaussian mixture, for which this integral is available in closed form \citep{kruger_predictive_inference_2021}. We compute the mean CRPS across the test set as
\begin{equation}
\mathrm{CRPS} = \frac1N\sum_n\Bigg[\frac{1}{M}\sum_{s\in\mathcal{M}} A\big(y_n-\mu_n^{(s)},\,\sigma^{(s)}\big)
- \frac{1}{2M^2}\sum_{s,s'\in\mathcal{M}} A\Big(\mu_n^{(s)}-\mu_n^{(s')},\,\sqrt{(\sigma^{(s)})^2+(\sigma^{(s')})^2}\Big)\Bigg]
\end{equation}
where $A(\mu,\sigma)=\mu\big(2\Phi(\mu/\sigma)-1\big)+2\sigma\varphi(\mu/\sigma)$, and $\mathcal{M}\subset\{1,\dots,S\}$ is a fixed uniform subsample of size $M=256$, drawn once with a common seed so all samplers are scored on the same draws. We compute the coverage as the share of test responses that fall inside the central $90\%$ interval of the posterior predictive. With the predictive CDF at $x_n$, $F_n(y) = \frac1S\sum_{s} \Phi\big((y-\mu_n^{(s)})/\sigma^{(s)}\big)$, the interval runs from $F_n^{-1}(0.05)$ to $F_n^{-1}(0.95)$, so we compute
\begin{equation}
\mathrm{C_{0.9}} = \frac1N\sum_{n=1}^N \mathbf 1\big\{0.05 \le F_n(y_n) \le 0.95\big\} ,
\end{equation}
from all $S$ draws. Values above $0.9$ indicate intervals that are too wide, and values below intervals that are too narrow.

For the UCI data we report the mean and standard error over the five splits. The one-dimensional and image experiments are single runs. For the image models, we average the class probabilities over $500$ posterior draws and report the accuracy, the NLL of the averaged probabilities, the expected calibration error (ECE) with $15$ equal-width confidence bins, and the Brier score.

\subsection{Baselines}
\label{sec:supp_baselines}
We here add additional information on the NUTS and stochastic gradient descent (SGD) baselines, and present the LBBNN method in more depth. NUTS is run with NumPyro \citep{phan2019composable} on the same priors and likelihood as the dense PDMPs, with a target acceptance probability of $0.9$ and the default maximum tree depth of $10$, using four chains of $1000$ warmup and $1000$ retained draws. For the image models, we also report a standard non-Bayesian network trained by stochastic gradient descent (SGD). It is a dense network trained by maximum likelihood, without the prior, and is fitted separately from the MAP $\beta_{\mathrm{ref}}$, which is described in \Cref{sec:supp_map}. SGD uses Nesterov momentum $0.9$, batch size $128$ and a learning rate of $0.05$, annealed by a cosine schedule over $40$ epochs.

\subsubsection{LBBNN method}
\label{sec:supp_lbnnn}
As a variational analogy to the sticky samplers, we have implemented the latent binary Bayesian neural network (LBBNN) from \citet{hubin_sparse_bnn}, based on their Git repository. In the LBBNN, one augments each weight with a binary inclusion variable $\gamma_{\ell, ij} \in \{0, 1\}$, which is then multiplied with the weight matrix, setting parameters to exactly zero. This is equivalent to placing a spike-and-slab prior on the weights. Bias terms carry no inclusion variable, analogous to biases being exempt from freezing in the sticky PDMPs.

The joint prior over layers factorizes as
\begin{equation}
    p(\boldsymbol{\beta}, \boldsymbol{\gamma})
    = \prod_{\ell=1}^{L} \Big[ \prod_{i,j} p(\gamma_{\ell,ij})\,
      \mathcal{N}\big((W_\ell)_{ij}; 0, \sigma_\ell^2\big)
      \prod_{k} \mathcal{N}\big((b_\ell)_k; 0, \sigma_b^2\big)\Big] \ ,
\end{equation}
where $\gamma_{\ell,ij} \sim \mathrm{Bernoulli}(w)$, with the same prior inclusion probability $w$ as the sticky samplers, in place of the Beta-Binomial prior used by the authors. The slab variance $\sigma_\ell^2$ and bias variance $\sigma_b^2$ are those of \Cref{sec:supp_model_specifications}, so the LBBNN shares the spike-and-slab prior of the sticky PDMPs. Inference for the LBBNN is performed via the variational family
\begin{equation}
    q(\boldsymbol{\beta}, \boldsymbol{\gamma})
    = \prod_{\ell,i,j}
      \mathcal{N}\big((W_\ell)_{ij}; \mu_{\ell,ij}, s_{\ell,ij}^2\big)\,
      \mathrm{Bernoulli}\big(\gamma_{\ell,ij}; \alpha_{\ell,ij}\big)
      \prod_{\ell,k} \mathcal{N}\big((b_\ell)_k; \mu^b_{\ell,k}, (s^b_{\ell,k})^2\big) \ ,
\end{equation}
to maximize the evidence lower bound
\begin{equation}
    \mathcal{L}
    = \mathbb{E}_{q}\big[\log p(\mathcal{D} \mid \boldsymbol{\beta}, \boldsymbol{\gamma})\big]
      - \mathrm{KL}\big(q(\boldsymbol{\beta}, \boldsymbol{\gamma}) \,\|\,
      p(\boldsymbol{\beta}, \boldsymbol{\gamma})\big) \ .
\end{equation}
Unlike \citet{hubin_sparse_bnn}, who fix the noise variance and tune it externally, we learn it on the UCI data as a point estimate under the same half-normal prior as the samplers, which puts the LBBNN on the same model as the samplers and requires no external tuning. During training the binary indicators $\boldsymbol{\gamma}$ are sampled by the continuous relaxation of \citet{hubin_sparse_bnn} with temperature $T_{\mathrm{temp}}$, tuned briefly to $T_{\mathrm{temp}}=0.5$ for the one-dimensional data and $T_{\mathrm{temp}}=0.2$ for the UCI data. During posterior draws we sample $\gamma_{\ell,ij} \sim \mathrm{Bern}(\alpha_{\ell,ij})$, so that weights can become exactly zero. We fit the models with Adam, using a single Monte Carlo sample per gradient evaluation, and the full dataset. For the one-dimensional data, the inclusion logits are drawn as $\lambda_{\ell,ij} \sim \mathcal U(0, 1)$, with $\alpha_{\ell,ij} = \operatorname{sigmoid}(\lambda_{\ell,ij})$, and the learning rate is fixed at $10^{-2}$ over $1.5\times 10^4$ epochs. For the UCI data, the inclusion probabilities are instead initialized close to one, with $\lambda_{\ell,ij} \sim \mathcal U(3.5, 4.5)$, and the learning rate decays from $10^{-2}$ to $10^{-4}$ by a cosine schedule over $5\times 10^4$ epochs. The variational means of the weights and biases start at the same MAP as the samplers, as does the noise level on the UCI data, while the slab standard deviations start small. These settings were chosen from brief tuning runs on two splits of the energy and yacht datasets, without the MAP start.

The LBBNN was developed with a Beta-Binomial prior and settings that differ from ours, so we also fit it with the noise variance fixed at a tuned value $\tau$ rather than learned. For each split, 20\% of the training data is held out for validation. Then $\tau$ is chosen, together with the initialisation of the inclusion probabilities, by the validation NLL among eight candidates, given by the residual variance of a least-squares linear fit scaled by factors between $0.005$ and $1$. The chosen model, started from random means, is refitted on all the training data of the split. Compared with the LBBNN started at the MAP, this tuning lowers the RMSE on every dataset for the medium network, but for the small network only on energy, and the RMSE remains behind the sticky samplers throughout. As the noise variance is then fixed in advance rather than learned, this LBBNN targets a different model than the samplers, and the main text reports the version with a learned noise variance.

\subsection{The bounding strategy}
\label{sec:supp_bounds}
After every window we set $t_{\max} \leftarrow t_{\max}\,\alpha^{\mathbb{I}\{\text{empty}\} - r}$, where $r$ is the number of rejected proposals and a window is empty when it reached its full length without an event \citep[Section 3.4]{andral2024}. We use $\alpha = 1.01$ for the one-dimensional and UCI experiments, and $\alpha = 1.02$ for the image models. The window then settles where empty windows and rejected proposals are equally frequent, balancing the two ways a window can spend gradient evaluations without producing an event. The tangent bound is inflated by a factor $1 + \varepsilon$ with $\varepsilon = 0.01$ for numerical slack.

When a violation $\lambda(\tau) > \tilde{\lambda}$ is detected at a proposed time $\tau$, the rest of the window is replaced by a rebuilt window of half the remaining width starting at $\tau$, and $t_{\max}$ is shrunk by a factor 2, or 1.1 for the image models, similar to \citet{andral2024}.

\subsection{Computing infrastructure}
\label{sec:supp_compute}
Experiments ran on an Apple M3 laptop (8 cores, 16GB), used for the one-dimensional and small UCI runs, and on a server with two Intel Xeon Gold 6226R CPUs (64 threads, 753GB RAM) and four NVIDIA A10 GPUs (24GB each). The medium UCI runs and NUTS used one CPU core per run, and the image models one GPU each. A one-dimensional PDMP run took about 6 minutes, a small UCI run about 30 minutes, a medium UCI run 2.5 to 4.5 hours and the MNIST models about 16 hours.

\section{Additional experiments}
\label{sec:supp_additional_experiments}
This section contains material that validates the ZigZag and Boomerang samplers on a banana-shaped target (\Cref{sec:supp_banana}) and the sticky samplers on a sparse linear regression against a Gibbs sampler (\Cref{sec:supp_sparse_linreg}). It also contains further one-dimensional examples (\Cref{sec:supp_extended_toys}), three more UCI datasets (\Cref{sec:supp_extended_UCI}), an ablation on the prior inclusion weight (\Cref{sec:supp_sparsity_ablation}), and details on the convergence analysis, the bounding scheme and minibatching.
\subsection{Banana distribution}
\label{sec:supp_banana}
\begin{figure}[ht]
\centering
\begin{subfigure}[b]{0.4\columnwidth}
    \centering
    \includegraphics[width=0.7\columnwidth]{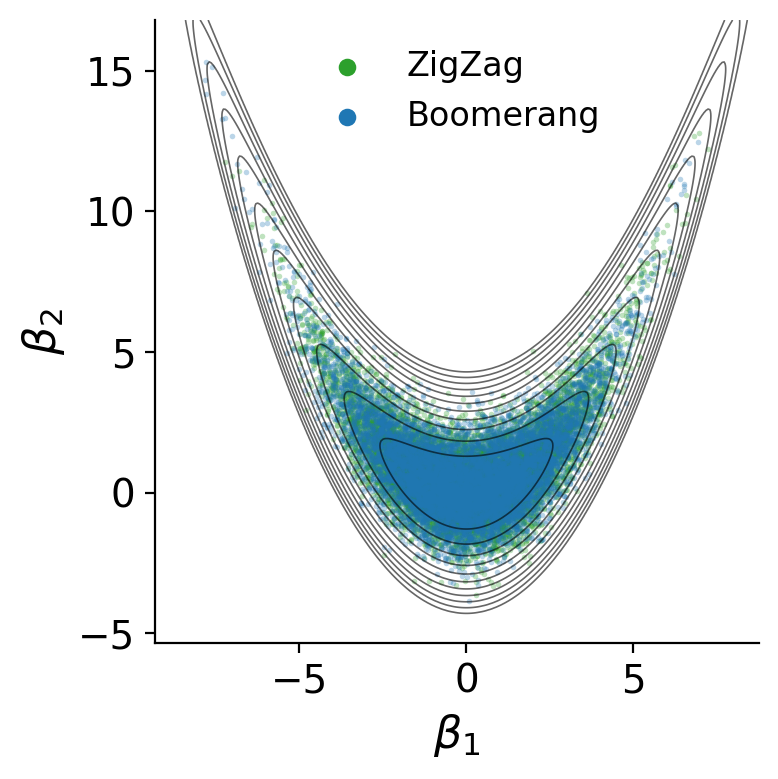}
    \label{fig:banana_contour}
\end{subfigure}
\hfill
\begin{subfigure}[b]{0.55\columnwidth}
    \centering
    \includegraphics[width=\columnwidth]{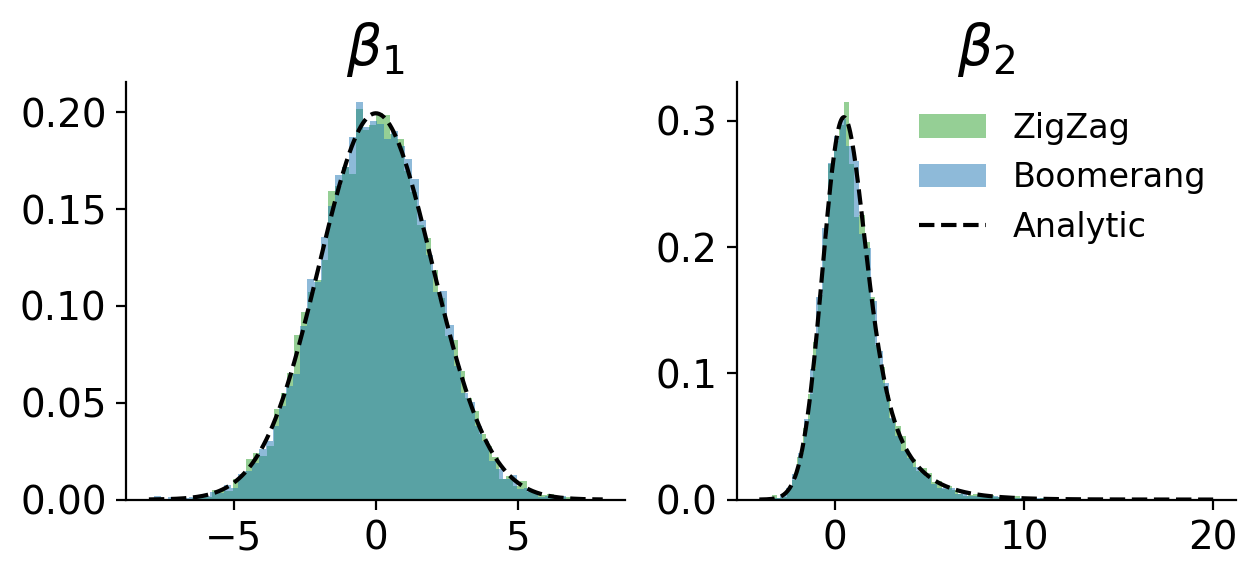}
    \label{fig:banana_marginals}
\end{subfigure}
\caption{ZigZag and Boomerang on the banana target. (a) Posterior draws overlaid on the target's density contours, and (b) the marginals against the exact analytic density.}
\label{fig:banana}
\end{figure}

As a validity check, we test our samplers on a 2-D banana-shaped target with a known closed-form density \citep{pagani2022}, and compare the resulting marginals and joint distribution against the analytic ground truth.
In \Cref{fig:banana} we sample from the banana target
\begin{equation}
E(\beta_1, \beta_2) = \tfrac{1}{2}(\beta_1/s)^2 + \tfrac{1}{2}\big(\beta_2 - a(\beta_1/s)^2\big)^2,
\end{equation}
with $a=1$, $s=2$, using the ZigZag and Boomerang samplers with $5\times 10^4$ events and $10^4$ draws after a burn-in of half the trajectory time. The Boomerang uses the reference $\mathcal{N}(\beta_{\mathrm{ref}}, \Sigma)$ with $\beta_{\mathrm{ref}} = (0, 1)$, $\Sigma^{-1} = \mathrm{diag}(1, 1/3)$ and refreshment rate $\lambda_{\mathrm{ref}} = 1$. As this is only two dimensions we have no reason to be sparse, so we do not apply the sticky samplers.

\begin{figure}[ht]
\centerline{\includegraphics[width=0.7\columnwidth]{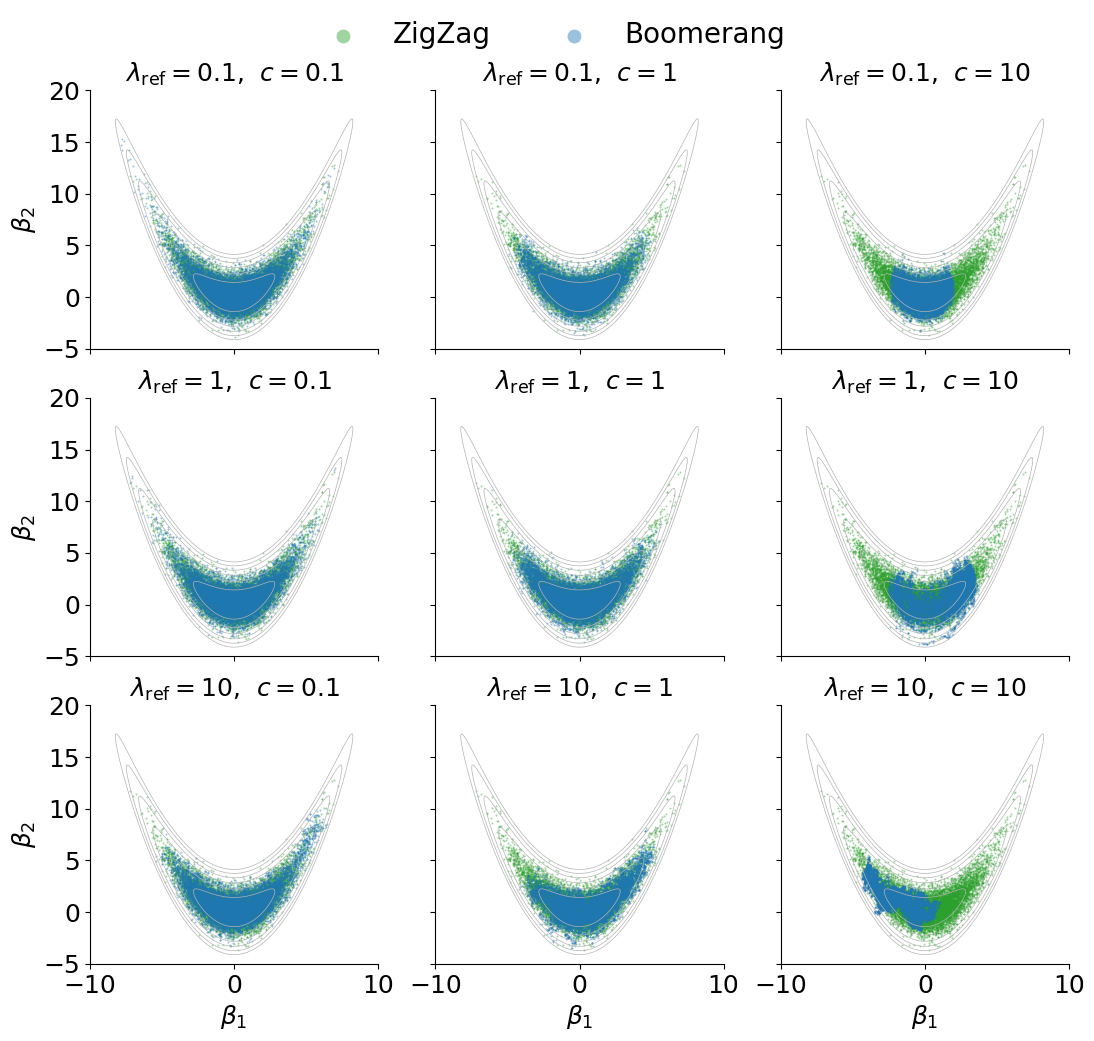}}
\caption{Boomerang (blue) on the banana target for reference precision scalings $c$ (columns) and refreshment rates $\lambda_{\mathrm{ref}}$ (rows), with the ZigZag (green) as a fixed benchmark in every panel.}
\label{fig:banana_sweep}
\end{figure}

The draws of both samplers cover the banana target, including its tails, and the sampled marginals are very similar and closely approximate the true distribution. This serves as a good indicator that the implementation is sound and that the samplers behave as expected. Furthermore, we test the sensitivity of the Boomerang with respect to its reference covariance and the refreshment rate. In \Cref{fig:banana_sweep}, we scale the reference precision by $c \in \{0.1, 1, 10\}$ and vary the refreshment rate $\lambda_{\mathrm{ref}} \in \{0.1, 1, 10\}$, giving nine settings with the ZigZag sampler as a fixed benchmark. With a wide reference ($c = 0.1$) the Boomerang covers the target as well as the ZigZag within the same budget, while a narrow reference ($c = 10$) confines it to the centre of the banana and misses the arms. The sampler is ergodic for every setting, but how well it explores within a fixed budget depends strongly on the reference measure. The refreshment rate plays a smaller role, with $\lambda_{\mathrm{ref}} = 10$ giving the least accurate results.

\subsection{Sparse linear regression}
\label{sec:supp_sparse_linreg}
We consider the model
\begin{equation}
    \mathbf{y} = \mathbf{X}\boldsymbol{\beta} + \varepsilon
\end{equation}
where $\mathbf{y} \in \mathbb{R}^n$ are the observations, $\mathbf{X} \in \mathbb{R}^{n \times d}$ are the covariates, $\boldsymbol{\beta} \in \mathbb{R}^d$ are the regression coefficients and $\varepsilon \in \mathbb{R}^n$ is i.i.d.\ Gaussian noise $\mathcal{N}(0, \sigma^2 \mathbf{I})$. In our experiment we treat $\sigma^2 = 1$ as known, set $n = 200$, $d = 20$, and generate covariates as $\mathbf{X} \sim \mathcal{N}(0, \mathbf{I})$. Furthermore, we let 5 of the coefficients be nonzero and geometrically decay them from $\beta_1 = 1.5$ with alternating signs. We place a spike-and-slab prior on the coefficients by introducing binary inclusion variables $\gamma_j$, $j = 1, \dots, d$,
\begin{equation}
    \label{eq:spike_slab}
    \gamma_j \overset{\text{iid}}{\sim} \mathrm{Bernoulli}(w),
    \qquad
    \beta_j \mid \gamma_j \sim
    \begin{cases}
        \mathcal{N}(0, \tau^2), & \gamma_j = 1, \\
        \delta_0,               & \gamma_j = 0,
    \end{cases}
\end{equation}
so that, marginally over $\gamma_j$,
\begin{equation}
    \label{eq:spike_slab_marginal}
    \beta_j \sim w\,\mathcal{N}(0, \tau^2) + (1 - w)\,\delta_0,
\end{equation}
which places positive mass on $\beta_j = 0$ and is the prior targeted by the sticky samplers, with the same prior inclusion weight $w$. The model is sampled via Gibbs sampling \citep{george1997approaches, ohara2009review}, exploiting conjugacy. Conditionally on $\boldsymbol{\gamma}$, the likelihood is $\mathbf{y} \mid \boldsymbol{\beta} \sim \mathcal{N}(\mathbf{X}\boldsymbol{\beta}, \sigma^2\mathbf{I})$, which is conjugate to the slab in \eqref{eq:spike_slab}, so $\boldsymbol{\beta}$
can be integrated out in closed form and
\begin{equation}
    \label{eq:factorization}
    p(\boldsymbol{\beta}, \boldsymbol{\gamma} \mid \mathbf{y})
    = p(\boldsymbol{\gamma} \mid \mathbf{y})\,
      p(\boldsymbol{\beta} \mid \boldsymbol{\gamma}, \mathbf{y}).
\end{equation}
Each Gibbs sweep updates the $\gamma_j$ one at a time from their Bernoulli full conditionals, after which $\boldsymbol{\beta}$ is drawn from its Gaussian conditional given $\boldsymbol{\gamma}$. As both conditionals are available exactly, we use the resulting draws as a reference posterior for the sticky
samplers. The same $\sigma^2$, $\tau^2$ and $w$ are used for the sticky samplers, so that all methods target the same posterior.

The sticky samplers start at the posterior mode $\hat{\boldsymbol{\beta}} = \mathbf{P}^{-1}\mathbf{X}^\top\mathbf{y}/\sigma^2$ of the dense model, with $\mathbf{P} = \mathbf{X}^\top\mathbf{X}/\sigma^2 + \tau^{-2}\mathbf{I}$, which also centres the diagonal Gaussian reference of the sticky Boomerang, with precision $\Sigma^{-1} = 0.1\mathrm{diag}(\mathbf{P})$, while Gibbs starts from the five covariates most correlated with $\mathbf{y}$. Both sticky samplers run for $P = 10^6$ events, and $S = 5\cdot 10^4$ draws are taken uniformly over the trajectory after a burn-in of $20\%$ of its time, while Gibbs runs $2000$ burn-in sweeps followed by $2 \cdot 10^4$ retained sweeps. We use $w = 0.2$ and $\tau = 1$, and repeat the experiment on five data sets generated with different seeds.

\begin{figure}[ht]
\centerline{\includegraphics[width=0.9\columnwidth]{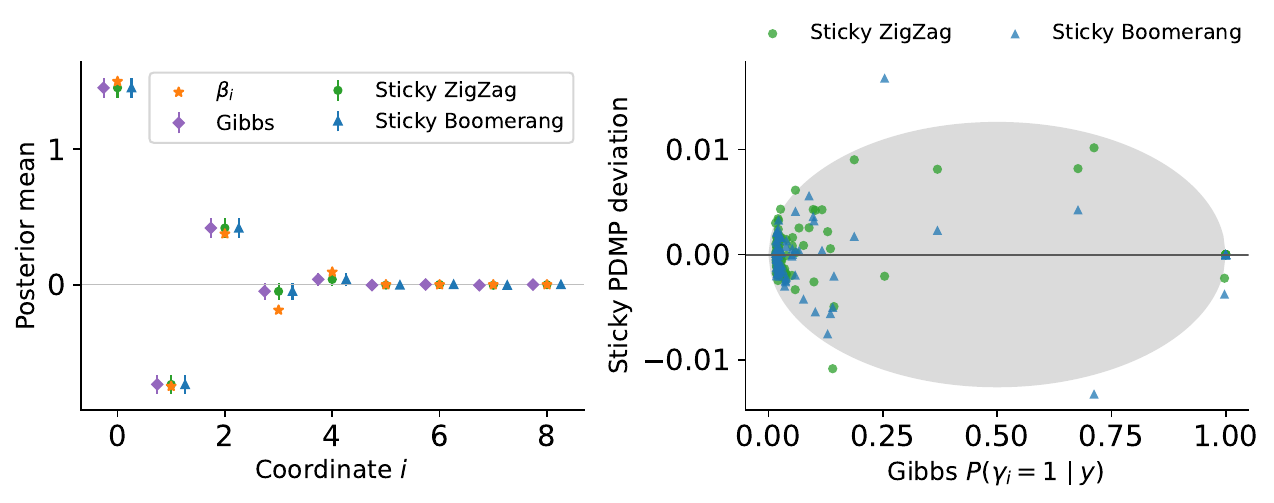}
}
\caption{Sparse linear regression, pooled over five data sets, with Gibbs as a reference sampler. Left, posterior means of the first nine coefficients with $\pm 1$ posterior standard deviation, and the true coefficients $\beta_i$. Right, deviation of the sticky PDMPs' posterior inclusion probabilities from those of Gibbs, with a band of two Monte Carlo standard deviations.}
\label{fig:linreg}
\end{figure}

The results are seen in \Cref{fig:linreg}, which pools the results from the five seeds. In the left panel, Gibbs and the sticky PDMPs agree closely on the posterior coefficient means, consistent with sampling the same target. The right panel shows the deviation of the sticky samplers' posterior inclusion probabilities from those of Gibbs. The deviations scatter symmetrically about zero with no trend in $P(\gamma_i = 1 \mid \mathbf{y})$, and lie mostly within the shaded band of two Monte Carlo standard deviations, computed from the median effective sample sizes of the Gibbs and PDMP inclusion indicators.

\subsection{More one-dimensional examples}
\label{sec:supp_extended_toys}
With $y_i \mid x_i \sim \mathcal{N}\big(f(x_i), \sigma^2\big)$, we fit the same models, using the same settings as in the main text, to three more one-dimensional datasets. The \emph{Hernandez} data follow \citet{Hernandez_Lobato_toy_bnn_2015}, with $f(x) = x^3$, $\sigma = 3$ and 20 points drawn uniformly from $[-4, 4]$. The \emph{multiscale} data combine a slow and a fast oscillation, with $f(x) = \sin(0.5x) + 0.3\sin(4x)$, $\sigma = 0.1$ and 80 points drawn uniformly from $[-3, 3]$. The \emph{gap} data have $f(x) = \sin(1.5x) + 0.3x$, $\sigma = 0.1$ and 30 points drawn uniformly from each of $[-3, -1.5]$ and $[1.5, 3]$, leaving a deliberate gap in $[-1.5, 1.5]$. The metrics in \Cref{tab:toys_extended} are computed on evenly spaced test points within the range of the training inputs, $[-4, 4]$ for the Hernandez data and $[-3, 3]$ otherwise, so for the gap data they include the gap.

\begin{figure}[ht]
\centering
\begin{minipage}{0.48\columnwidth}
\centering
\includegraphics[width=\columnwidth]{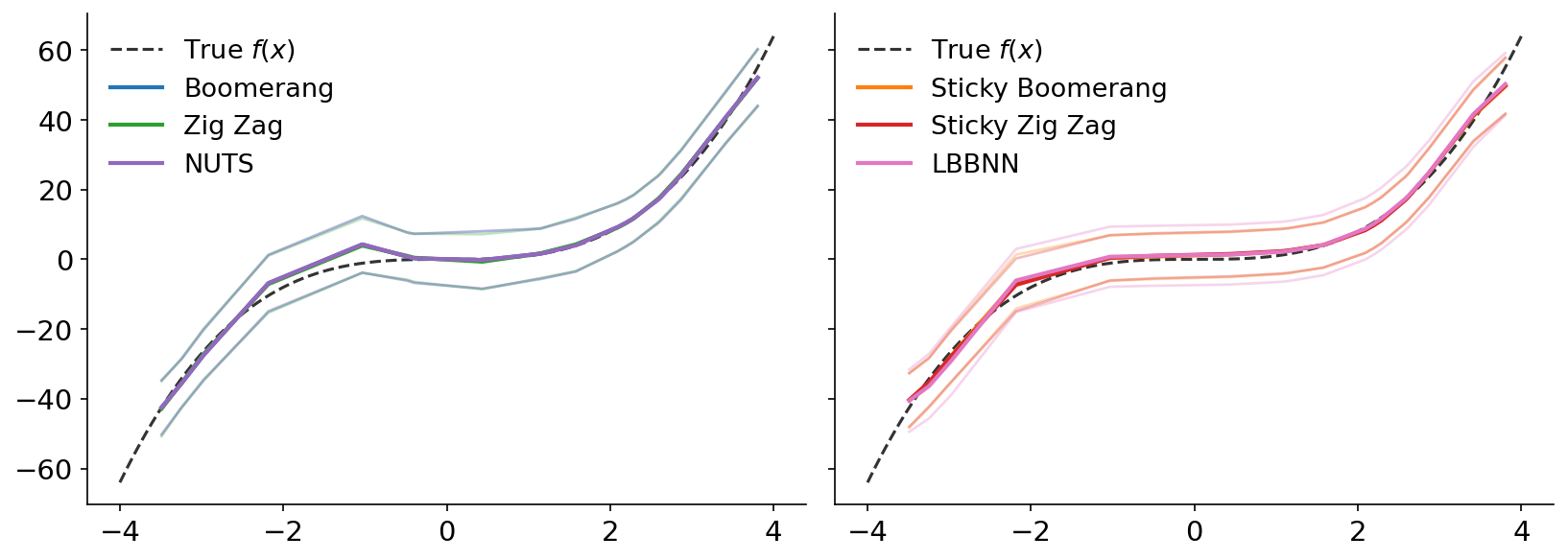}
\end{minipage}
\hfill
\begin{minipage}{0.48\columnwidth}
\centering
\includegraphics[width=\columnwidth]{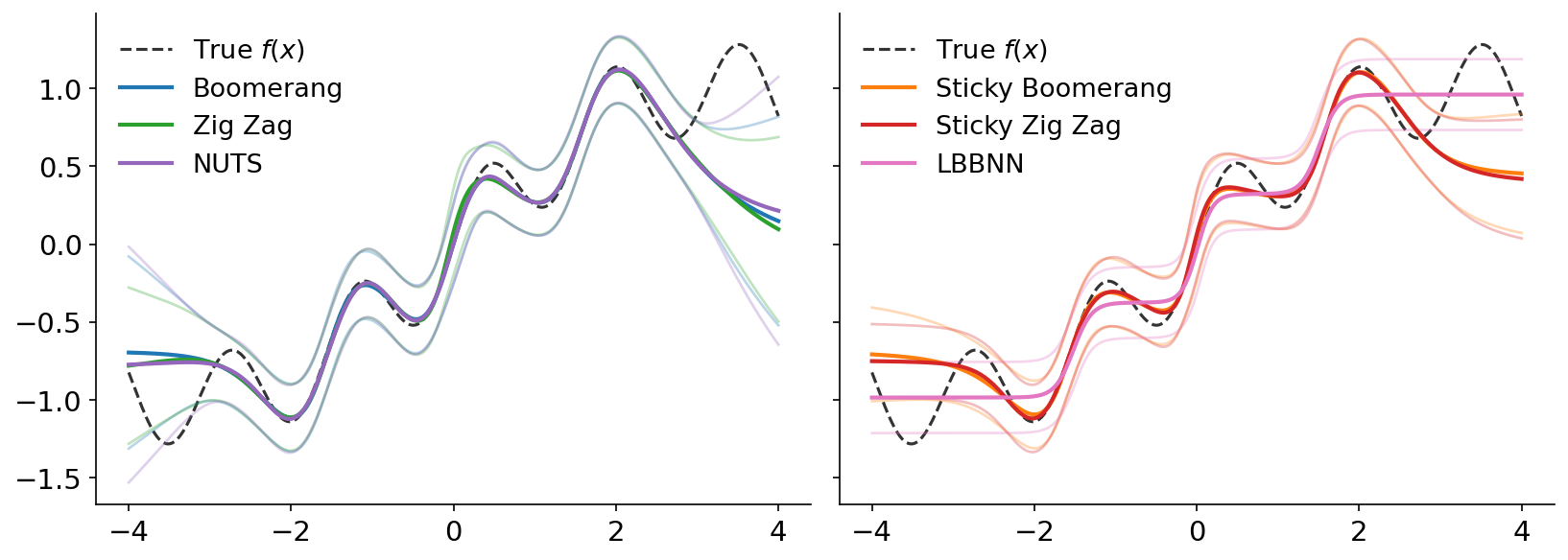}
\end{minipage}
\hfill
\begin{minipage}{0.6\columnwidth}
\centering
\includegraphics[width=\columnwidth]{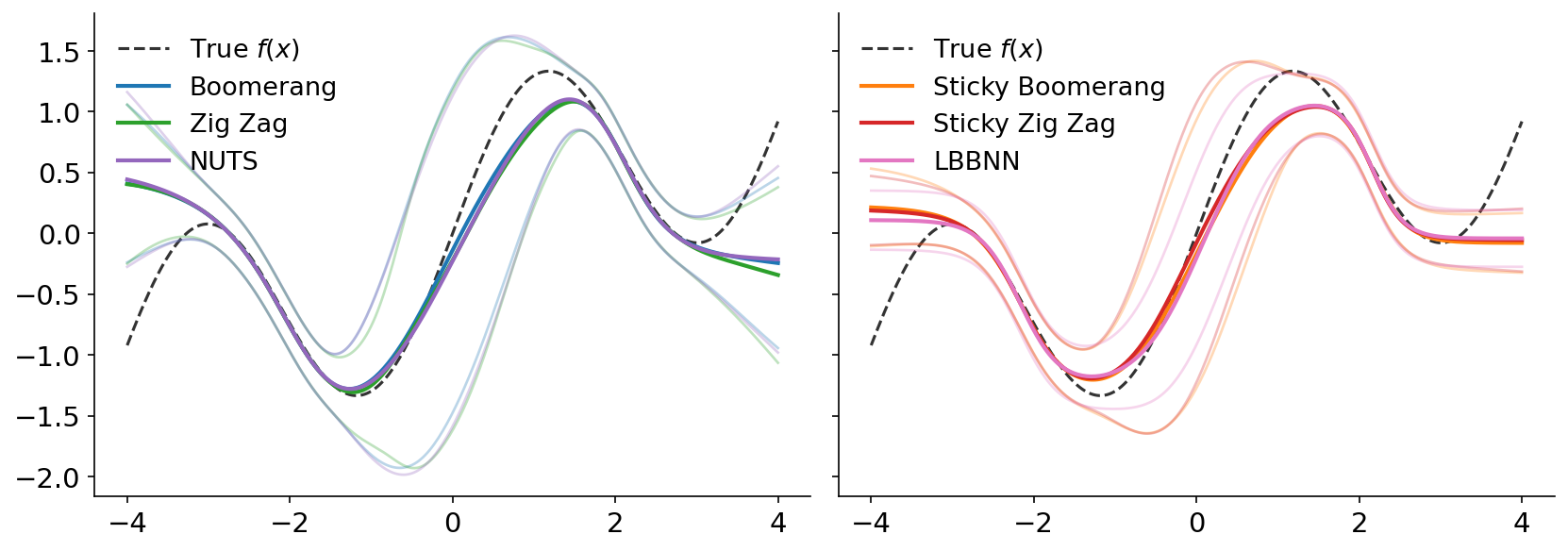}
\end{minipage}
\caption{Posterior predictive fits on the \emph{Hernandez} (top left), \emph{multiscale} (top right) and \emph{gap} (bottom) datasets. In each panel, the dashed black line is the true function, the solid coloured lines are the posterior predictive means, and the thin lines of the same colour lie two predictive standard deviations above and below the mean.}
\label{fig:toy_bnn}
\end{figure}

\begin{table}
\centering
\resizebox{\columnwidth}{!}{
\begin{tabular}{l cccc cccc cccc}
\toprule
& \multicolumn{4}{c}{Hernandez} & \multicolumn{4}{c}{Multiscale} & \multicolumn{4}{c}{Gap} \\
\cmidrule(lr){2-5} \cmidrule(lr){6-9} \cmidrule(lr){10-13}
Sampler & Sparsity & RMSE & NLL & CRPS & Sparsity & RMSE & NLL & CRPS & Sparsity & RMSE & NLL & CRPS \\
\midrule
Boomerang        & -- & 4.129 & 2.782 & 2.293 & -- & \textbf{0.119} & -0.710 & 0.067 & -- & 0.199 & \textbf{-0.277} & 0.118 \\
ZigZag           & -- & \textbf{3.989} & \textbf{2.762} & \textbf{2.134} & -- & \textbf{0.119} & -0.707 & \textbf{0.066} & -- & 0.224 & -0.167 & 0.128 \\
NUTS             & -- & 4.121 & 2.781 & 2.309 & -- & \textbf{0.119} & \textbf{-0.713} & 0.067 & -- & 0.214 & -0.275 & 0.116 \\
Sticky Boomerang & 87\% & 4.498 & 2.832 & 2.425 & 85\% & 0.127 & -0.626 & 0.071 & 85\% & 0.210 & -0.140 & 0.117 \\
Sticky ZigZag    & 87\% & 4.369 & 2.811 & 2.383 & 84\% & 0.123 & -0.665 & 0.069 & 86\% & \textbf{0.196} & -0.141 & \textbf{0.111} \\
LBBNN            & 95\% & 4.677 & 2.942 & 2.570 & 95\% & 0.177 & -0.084 & 0.105 & 89\% & 0.201 & -0.185 & \textbf{0.111} \\
\bottomrule
\end{tabular}
}
\caption{Predictive performance on the one-dimensional datasets.}
\label{tab:toys_extended}
\end{table}

The non-sticky PDMPs and NUTS again give nearly identical fits. On the Hernandez and multiscale data the dense samplers are slightly better than the sticky samplers, which are in turn better than the LBBNN, most clearly on the multiscale data. On the gap data all methods perform alike, with the sticky ZigZag lowest in RMSE, the sticky ZigZag and the LBBNN tied in CRPS, and the dense Boomerang and NUTS lowest in NLL.

\subsection{More UCI datasets}
\label{sec:supp_extended_UCI}
\Cref{tab:energy,tab:concrete,tab:yacht} show the results for the UCI datasets not included in the main text, with the same settings as for Boston. They show the same overall pattern. The sticky samplers are competitive with the non-sticky samplers and NUTS, with differences mostly within the between-split variation, and the LBBNN is clearly behind in RMSE and CRPS on every dataset. The cost of sparsity is most visible for the small network on concrete, where the RMSE of the sticky samplers grows from 5.15 to 5.64 as they become sparser.

Sparsity grows as $w$ decreases, but how many weights are set to zero also depends on the dataset and the network. For the medium network the sparsity stays close to the prior sparsity $1-w$, as for Boston. For the small network it is clearly lower on all three datasets, with 42 to 61\% at $w=0.3$ and 51 to 80\% at $w=0.1$.

For the medium network on concrete, both the non-sticky and the sticky Boomerang have a clearly higher NLL than the other samplers and cover only 80 to 81\% of the test responses, as for Boston. This matches the convergence analysis on Boston (\Cref{sec:convergence}), where the Boomerang samplers on the medium network stay close to the MAP they start from. For the medium network on energy, NUTS is clearly better than all PDMP samplers, while on yacht it is the weakest of the samplers for the medium network. We expect these differences to be a result of samplers not converging, although it is hard to be completely sure as the convergence analysis was only done for the Boston dataset. The LBBNN covers 89 to 96\% of the test responses, close to or above the nominal level despite its larger errors.

\begin{table}
\centering
\resizebox{\columnwidth}{!}{
\begin{tabular}{rllccccc}
\toprule
Network & Sampler & $w$ & Sparsity & RMSE & NLL & CRPS & $C_{0.9}$ \\
\midrule
$d=502$
& Boomerang        & --   & --   & 0.35 $\pm$ .02 & 0.37 $\pm$ .06 & 0.19 $\pm$ .01 & \textbf{90\%} \\
& ZigZag           & --   & --   & 0.34 $\pm$ .01 & 0.35 $\pm$ .03 & 0.19 $\pm$ .01 & 91\% \\
& NUTS             & --   & --   & \textbf{0.33 $\pm$ .02} & \textbf{0.30 $\pm$ .04} & \textbf{0.18 $\pm$ .01} & 91\% \\
& Sticky Boomerang & 0.3  & 42\% & 0.34 $\pm$ .02 & 0.35 $\pm$ .07 & 0.19 $\pm$ .01 & \textbf{90\%} \\
& Sticky ZigZag    & 0.3  & 50\% & 0.34 $\pm$ .01 & 0.35 $\pm$ .03 & 0.19 $\pm$ .01 & 91\% \\
& Sticky Boomerang & 0.1  & 51\% & 0.35 $\pm$ .02 & 0.39 $\pm$ .07 & 0.19 $\pm$ .01 & 88\% \\
& Sticky ZigZag    & 0.1  & 62\% & 0.35 $\pm$ .01 & 0.37 $\pm$ .03 & 0.19 $\pm$ .01 & 91\% \\
& LBBNN            & 0.3  & 80\% & 0.45 $\pm$ .01 & 0.67 $\pm$ .01 & 0.26 $\pm$ .01 & 94\% \\
\midrule
$d=5602$
& Boomerang        & --   & --   & 0.35 $\pm$ .02 & 0.37 $\pm$ .04 & 0.19 $\pm$ .01 & 93\% \\
& ZigZag           & --   & --   & 0.35 $\pm$ .02 & 0.36 $\pm$ .04 & 0.19 $\pm$ .01 & \textbf{92\%} \\
& NUTS             & --   & --   & \textbf{0.27 $\pm$ .02} & \textbf{0.04 $\pm$ .05} & \textbf{0.14 $\pm$ .01} & \textbf{92\%} \\
& Sticky Boomerang & 0.3  & 66\% & 0.36 $\pm$ .02 & 0.41 $\pm$ .04 & 0.20 $\pm$ .01 & \textbf{92\%} \\
& Sticky ZigZag    & 0.3  & 67\% & 0.34 $\pm$ .02 & 0.35 $\pm$ .03 & 0.19 $\pm$ .01 & 93\% \\
& Sticky Boomerang & 0.1  & 86\% & 0.36 $\pm$ .02 & 0.42 $\pm$ .04 & 0.20 $\pm$ .01 & \textbf{92\%} \\
& Sticky ZigZag    & 0.1  & 87\% & 0.36 $\pm$ .02 & 0.39 $\pm$ .04 & 0.20 $\pm$ .01 & \textbf{92\%} \\
& LBBNN            & 0.3  & 88\% & 0.57 $\pm$ .04 & 0.90 $\pm$ .06 & 0.32 $\pm$ .02 & 96\% \\
\bottomrule
\end{tabular}
}
\caption{
Predictive performance on the UCI energy dataset, mean $\pm$ SEM over 5 independent train/test splits. Sparsity denotes the percentage of weights sampled exactly zero, averaged over draws. $C_{0.9}$ is the share of test responses inside the central 90\% predictive interval, with an SEM of 1 to 3 percentage points.
}

\label{tab:energy}
\end{table}

\begin{table}
\centering
\resizebox{\columnwidth}{!}{
\begin{tabular}{rllccccc}
\toprule
Network & Sampler & $w$ & Sparsity & RMSE & NLL & CRPS & $C_{0.9}$ \\
\midrule
$d=502$
& Boomerang        & --   & --   & \textbf{4.97 $\pm$ .20} & \textbf{3.01 $\pm$ .04} & \textbf{2.70 $\pm$ .09} & 87\% \\
& ZigZag           & --   & --   & 5.12 $\pm$ .15 & 3.03 $\pm$ .03 & 2.79 $\pm$ .07 & 88\% \\
& NUTS             & --   & --   & 5.15 $\pm$ .19 & 3.03 $\pm$ .03 & 2.80 $\pm$ .08 & \textbf{89\%} \\
& Sticky Boomerang & 0.3  & 54\% & 5.15 $\pm$ .22 & 3.05 $\pm$ .04 & 2.81 $\pm$ .09 & 88\% \\
& Sticky ZigZag    & 0.3  & 61\% & 5.38 $\pm$ .20 & 3.09 $\pm$ .04 & 2.96 $\pm$ .10 & 88\% \\
& Sticky Boomerang & 0.1  & 72\% & 5.32 $\pm$ .24 & 3.08 $\pm$ .04 & 2.91 $\pm$ .10 & 87\% \\
& Sticky ZigZag    & 0.1  & 80\% & 5.64 $\pm$ .18 & 3.14 $\pm$ .03 & 3.09 $\pm$ .08 & 88\% \\
& LBBNN            & 0.3  & 82\% & 6.26 $\pm$ .19 & 3.25 $\pm$ .03 & 3.47 $\pm$ .10 & \textbf{89\%} \\
\midrule
$d=5602$
& Boomerang        & --   & --   & 4.89 $\pm$ .31 & 3.45 $\pm$ .18 & \textbf{2.46 $\pm$ .11} & 81\% \\
& ZigZag           & --   & --   & 4.79 $\pm$ .28 & 2.97 $\pm$ .07 & 2.49 $\pm$ .12 & 87\% \\
& NUTS             & --   & --   & 4.88 $\pm$ .27 & \textbf{2.94 $\pm$ .05} & 2.56 $\pm$ .11 & 88\% \\
& Sticky Boomerang & 0.3  & 75\% & 4.90 $\pm$ .31 & 3.51 $\pm$ .19 & 2.47 $\pm$ .11 & 81\% \\
& Sticky ZigZag    & 0.3  & 68\% & \textbf{4.76 $\pm$ .25} & 2.97 $\pm$ .06 & 2.52 $\pm$ .12 & 89\% \\
& Sticky Boomerang & 0.1  & 85\% & 4.91 $\pm$ .30 & 3.50 $\pm$ .19 & 2.48 $\pm$ .11 & 80\% \\
& Sticky ZigZag    & 0.1  & 88\% & 4.85 $\pm$ .16 & 2.97 $\pm$ .03 & 2.60 $\pm$ .08 & 87\% \\
& LBBNN           & 0.3  & 88\% & 6.42 $\pm$ .17 & 3.28 $\pm$ .02 & 3.58 $\pm$ .08 & \textbf{90\%} \\
\bottomrule
\end{tabular}
}
\caption{
Predictive performance on the UCI concrete dataset, mean $\pm$ SEM over 5 independent train/test splits. Sparsity denotes the percentage of weights sampled exactly zero, averaged over draws. $C_{0.9}$ is the share of test responses inside the central 90\% predictive interval, with an SEM of 0 to 2 percentage points.
}
\label{tab:concrete}
\end{table}

\begin{table}
\centering
\resizebox{\columnwidth}{!}{
\begin{tabular}{rllccccc}
\toprule
Network & Sampler & $w$ & Sparsity & RMSE & NLL & CRPS & $C_{0.9}$ \\
\midrule
$d=402$
& Boomerang        & --   & --   & 0.32 $\pm$ .07 & 0.45 $\pm$ .26 & 0.17 $\pm$ .02 & 92\% \\
& ZigZag           & --   & --   & 0.33 $\pm$ .07 & 0.68 $\pm$ .40 & 0.15 $\pm$ .02 & 87\% \\
& NUTS             & --   & --   & \textbf{0.28 $\pm$ .08} & \textbf{0.15 $\pm$ .62} & \textbf{0.11 $\pm$ .02} & 88\% \\
& Sticky Boomerang & 0.3  & 54\% & 0.34 $\pm$ .07 & 0.84 $\pm$ .46 & 0.16 $\pm$ .02 & 85\% \\
& Sticky ZigZag    & 0.3  & 56\% & 0.33 $\pm$ .06 & 0.63 $\pm$ .35 & 0.15 $\pm$ .02 & 86\% \\
& Sticky Boomerang & 0.1  & 67\% & 0.32 $\pm$ .06 & 0.56 $\pm$ .36 & 0.15 $\pm$ .02 & 85\% \\
& Sticky ZigZag    & 0.1  & 70\% & 0.36 $\pm$ .08 & 0.75 $\pm$ .40 & 0.16 $\pm$ .02 & \textbf{90\%} \\
& LBBNN            & 0.3  & 85\% & 0.73 $\pm$ .09 & 0.95 $\pm$ .06 & 0.35 $\pm$ .02 & 92\% \\
\midrule
$d=5502$
& Boomerang        & --   & --   & 0.30 $\pm$ .05 & \textbf{-0.37 $\pm$ .13} & \textbf{0.12 $\pm$ .02} & 88\% \\
& ZigZag           & --   & --   & 0.29 $\pm$ .05 & -0.07 $\pm$ .16 & 0.13 $\pm$ .02 & 93\% \\
& NUTS             & --   & --   & 0.41 $\pm$ .10 & 0.29 $\pm$ .35 & 0.15 $\pm$ .03 & 79\%  \\
& Sticky Boomerang & 0.3  & 67\% & \textbf{0.28 $\pm$ .04} & -0.08 $\pm$ .16 & 0.13 $\pm$ .02 & \textbf{90\%} \\
& Sticky ZigZag    & 0.3  & 68\% & 0.32 $\pm$ .06 & 0.29 $\pm$ .29 & 0.15 $\pm$ .03 & 86\%  \\
& Sticky Boomerang & 0.1  & 87\% & 0.30 $\pm$ .04 & 0.03 $\pm$ .16 & 0.14 $\pm$ .02 & 91\%  \\
& Sticky ZigZag    & 0.1  & 87\% & 0.30 $\pm$ .04 & 0.14 $\pm$ .17 & 0.14 $\pm$ .02 & 85\%  \\
& LBBNN            & 0.3  & 88\% & 0.86 $\pm$ .10 & 1.25 $\pm$ .02 & 0.45 $\pm$ .02 & 96\% \\
\bottomrule
\end{tabular}
}
\caption{
Predictive performance on the UCI yacht dataset, mean $\pm$ SEM over 5 independent train/test splits. Sparsity denotes the percentage of weights sampled exactly zero, averaged over draws. $C_{0.9}$ is the share of test responses inside the central 90\% predictive interval, with an SEM of 1 to 5 percentage points.
}
\label{tab:yacht}
\end{table}

\subsection{More convergence diagnostics}
\label{sec:supp_convergence}
\Cref{tab:supp_rhat} reports the rank-normalised $\hat R$ and bulk effective sample size (ESS) of \citet{Vehtari_rank_norm_rhat} for the convergence runs of \Cref{sec:convergence}, computed with ArviZ \citep{Martin2026}. Following \citet{Vehtari_rank_norm_rhat}, we take $\hat R \le 1.01$ and a bulk ESS above 400 as signs of convergence. For the PDMPs, the ESS is computed from the $S$ draws of each chain, ordered in time. This is not the effective sample size of the continuous-time average over the trajectory, as defined for the ZigZag by \citet[Section 5]{Bierkens_limits_2017}, but of the draws we use for prediction. The last two columns divide the ESS by the gradient evaluations spent, which puts the PDMPs and NUTS on the same budget per gradient evaluation, leaving out the Hessian-vector products of the bounds.

\begin{table}
\centering
\resizebox{\columnwidth}{!}{
\begin{tabular}{llccccccc}
\toprule
 & & & \multicolumn{2}{c}{$\hat R$} & \multicolumn{2}{c}{bulk ESS} & \multicolumn{2}{c}{bulk ESS per $10^7$ grad.\ evals} \\
\cmidrule(lr){4-5} \cmidrule(lr){6-7} \cmidrule(lr){8-9}
Network & Sampler & $w$ & $\log\sigma$ & test RMSE & $\log\sigma$ & test RMSE & $\log\sigma$ & test RMSE \\
\midrule
$d=752$
 & ZigZag & -- & 1.00 & 1.00 & 2{,}362 & 3{,}799 & 591 & 950 \\
 & Boomerang & -- & 1.01 & 1.00 & 2{,}622 & 9{,}629 & 656 & 2{,}407 \\
 & NUTS & -- & 1.00 & 1.00 & 4{,}854 & 3{,}961 & 12{,}965 & 10{,}580 \\
 & Sticky ZigZag & 0.3 & 1.00 & 1.00 & 2{,}763 & 3{,}529 & 691 & 882 \\
 & Sticky Boomerang & 0.3 & 1.01 & 1.00 & 1{,}676 & 4{,}095 & 419 & 1{,}024 \\
 & Sticky ZigZag & 0.1 & 1.00 & 1.00 & 2{,}236 & 2{,}385 & 559 & 596 \\
 & Sticky Boomerang & 0.1 & 1.04 & 1.00 & 62 & 2{,}151 & 16 & 538 \\
\midrule
$d=5852$
 & ZigZag & -- & 1.03 & 1.01 & 173 & 547 & 43 & 137 \\
 & Boomerang & -- & 1.63 & 1.73 & 6.5 & 6.1 & 1.6 & 1.5 \\
 & NUTS & -- & 1.00 & 1.00 & 1{,}624 & 3{,}478 & 2{,}276 & 4{,}875 \\
 & Sticky ZigZag & 0.3 & 1.04 & 1.04 & 67 & 191 & 17 & 48 \\
 & Sticky Boomerang & 0.3 & 1.70 & 2.03 & 6.2 & 5.3 & 1.5 & 1.3 \\
 & Sticky ZigZag & 0.1 & 1.03 & 1.01 & 109 & 294 & 27 & 74 \\
 & Sticky Boomerang & 0.1 & 1.79 & 2.00 & 5.9 & 5.4 & 1.5 & 1.3 \\
\bottomrule
\end{tabular}
}
\caption{Convergence diagnostics on the first Boston split, four chains per sampler started from different MAPs, with $10^7$
gradient evaluations per PDMP chain and one NUTS chain per MAP. Rank-normalised $\hat R$ and bulk ESS of $\log \sigma$
and the test RMSE, pooled over the four chains, which hold $16{,}000$ draws for the PDMPs and 4000 for NUTS.
The last two columns are rates, the bulk ESS per $10^7$ gradient evaluations, the budget of one PDMP chain. For NUTS, warmup is included when counting gradients.
}
\label{tab:supp_rhat}
\end{table}

\begin{figure}[ht]
\centering
\begin{minipage}{0.99\columnwidth}
\centering
\includegraphics[width=\columnwidth]{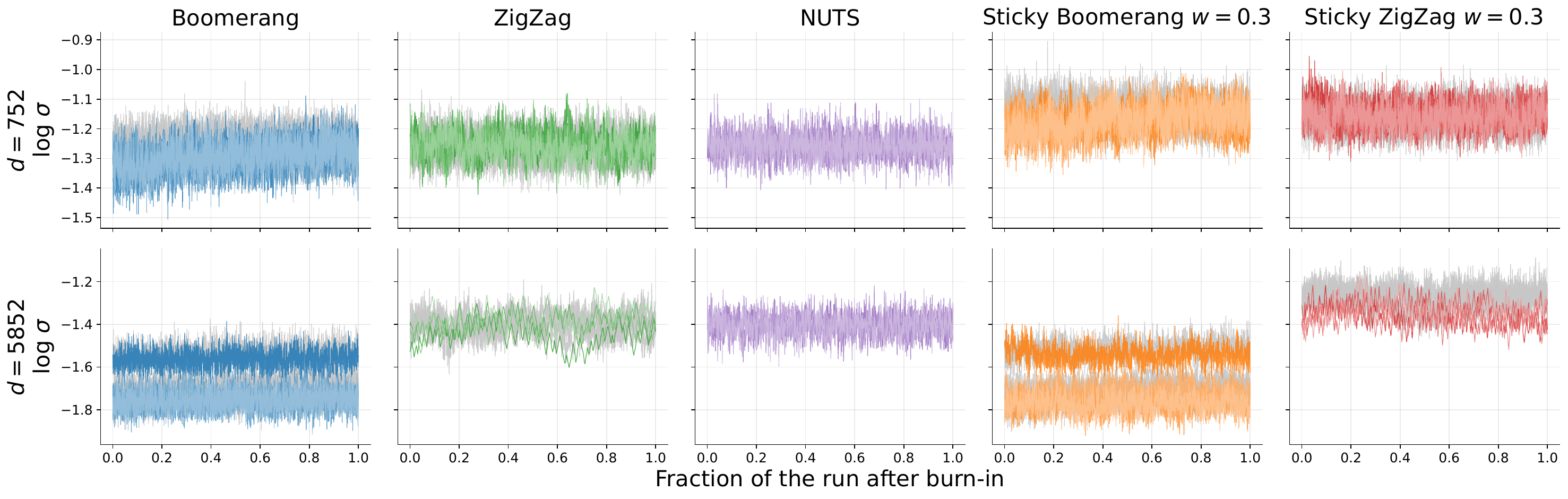}
\end{minipage}
\hfill
\begin{minipage}{0.99\columnwidth}
\centering
\includegraphics[width=\columnwidth]{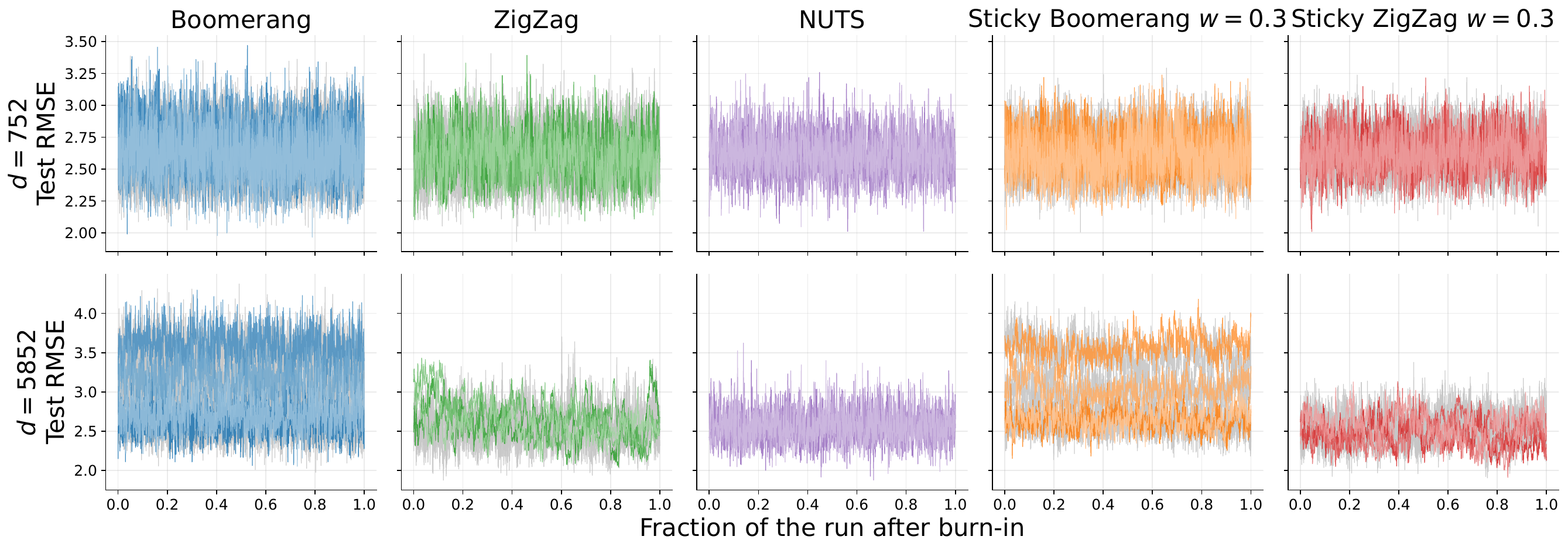}
\end{minipage}
\hfill
\begin{minipage}{0.99\columnwidth}
\centering
\includegraphics[width=\columnwidth]{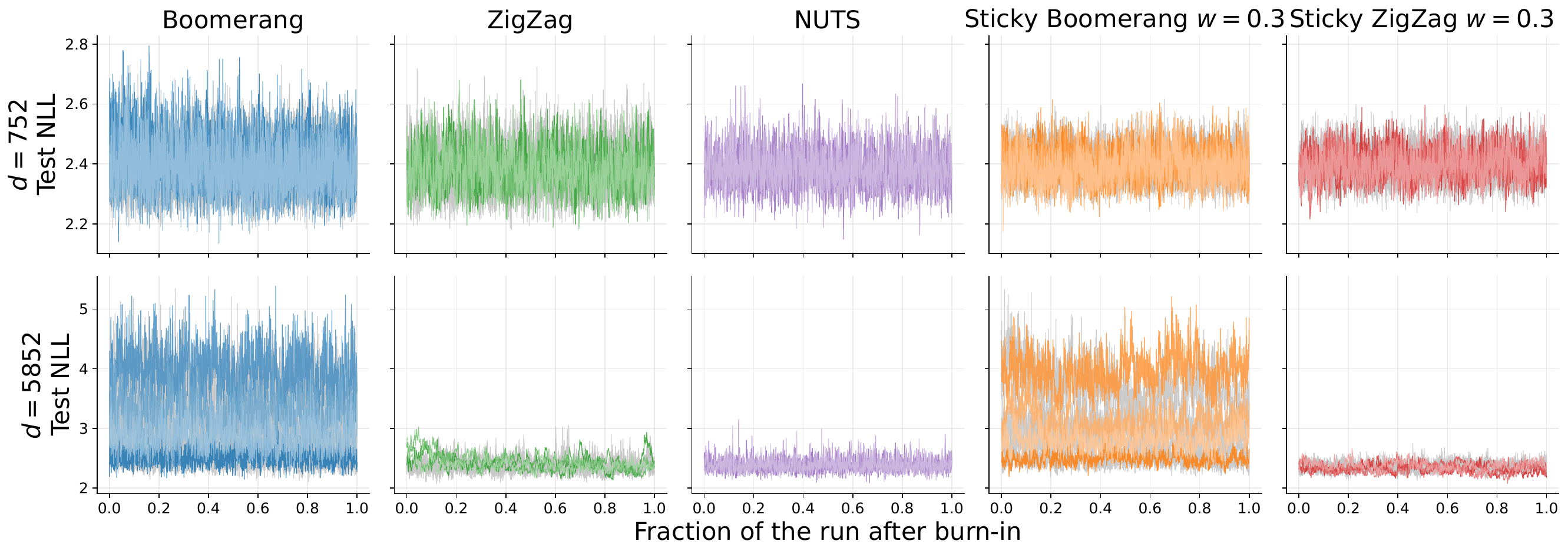}
\end{minipage}
\caption{Traces after burn-in on the first Boston split, with $G=10^6$ (coloured) and $G=10^7$ (grey) gradient evaluations per chain. The NLL is per draw, not of the predictive mixture.}

\label{fig:traces}
\end{figure}

\subsection{Effect of prior inclusion weight}
\label{sec:supp_sparsity_ablation}
Sampling from sticky PDMPs with sticking parameters $\kappa_i$ targets the posterior arising from a spike and slab prior \citep{bierkens2022sticky}. The parameters $\kappa_i$ play a central part of the modelling and are assumed known a priori. In our experiments the slab is Gaussian $\mathcal{N}(0, \sigma^2_{\ell})$ with the variance scaled by the fan-in $\sigma_{\ell} = \frac{\sigma_w}{\sqrt{F_{\ell}}}$
This gives the relation to $\kappa_i$ from \citet{bierkens2022sticky} as
\begin{equation}
\kappa_i = \frac{w}{1-w}\cdot\frac{1}{\sqrt{2\pi}\sigma_{\ell}}
\end{equation}
where $w \in (0, 1)$ is the prior inclusion probability, shared across all weights in the network. Since $w$ has a more direct interpretation as the prior probability that a parameter is included in the model, we focus our sparsity ablation study on the effect $w$ has on the sticky PDMPs.

\begin{figure}[ht]
\centerline{\includegraphics[width=0.8\columnwidth]{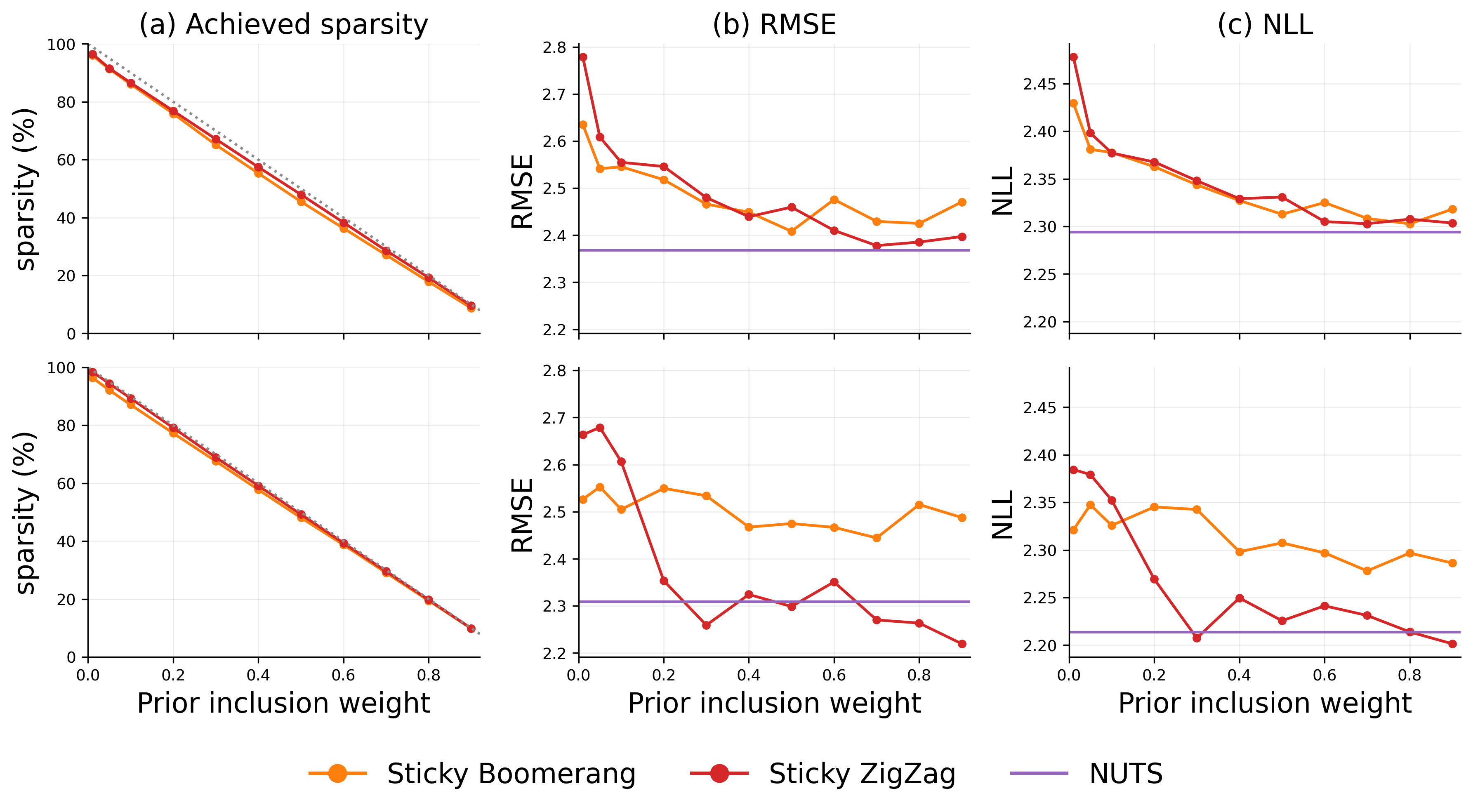}}
\caption{Effect of the prior inclusion weight $w$ on Boston for the small ($d = 752$, top) and medium ($d = 5852$, bottom) networks. (a) Share of weights that are exactly zero, averaged over the posterior draws, with the prior sparsity $1-w$ dotted. (b) Test RMSE and (c) test NLL, with dense NUTS as a horizontal reference.}
\label{fig:sparsity_ablation}
\end{figure}

In \Cref{fig:sparsity_ablation} we vary $w$ from $0.01$ to $0.9$ for both sticky samplers on the first split of the Boston dataset, for the small and the medium network, and report the achieved sparsity together with the test RMSE and NLL. Every run uses $10^6$ gradient evaluations and starts from the same MAP, and dense NUTS serves as a reference. On Boston, the achieved sparsity, measured as the share of weights that are exactly zero averaged over the posterior draws, closely follows the prior sparsity $1-w$ for both networks and both samplers (\Cref{fig:sparsity_ablation}a), which makes $w$ a direct handle on the size of the model. This is not the case for every dataset. For the small network on energy, concrete and yacht, the samplers keep clearly more weights than $1-w$ implies (\Cref{sec:supp_extended_UCI}), so the data can override the prior. Still, $w$ must be chosen in advance, and it largely decides how sparse the model becomes.

In the small network, predictive performance improves as $w$ grows and levels off from about $w = 0.5$, where the sticky samplers follow NUTS closely, while very sparse models lose accuracy (\Cref{fig:sparsity_ablation}b,c). In the larger network, the sticky ZigZag shows the same pattern, although the curves fluctuate more between neighbouring values of $w$. The sticky Boomerang varies little across $w$ and stays above NUTS throughout. These curves should also be read with the convergence analysis of \Cref{sec:convergence} and \Cref{sec:supp_convergence} in mind. On the small network, the test RMSE and NLL of all samplers stabilise early (\Cref{fig:traces}), so we take these curves at face value. For the medium network, the same holds for the trend of the ZigZag, but we interpret the Boomerang's flat curve as a result of slow mixing rather than an effect of $w$.

\subsection{Comparison of bounding strategies}
\label{sec:supp_grid_vs_singlesegment}

In~\citet{andral2024}, an interval $[0,t_{\max}]$ is divided into $K$ subsegments, and a piecewise constant upper bound is found on the rate function $\lambda(t)$. In this section, we compare this with the approach used throughout this paper, where the whole interval is bounded by a single constant, $K = 1$. We demonstrate that fewer gradient evaluations are then needed, as the rate is no longer evaluated at grid points beyond the proposed event time.

In our implementation of the \citet{andral2024} procedure, $K = \min(\max(\lceil t_{\max}/\Delta \rceil, 2), K_{\max})$ for a grid spacing $\Delta$, and $t_{\max}$ was adapted similarly to \citet{andral2024}, divided by $1.04$ after a window with any rejection and otherwise multiplied by $1.01$ after an empty window. Our scheme uses $K=1$, with $t_{\max}$ adapted similarly.

To compare the two schemes, consider a constant rate $\lambda$ as a simple model of the cost. The bounds from both bounding schemes are then exact up to the inflation $\varepsilon$, accepted proposals arrive at rate $\lambda$ and rejections at rate $\varepsilon\lambda$. The time of the first accepted proposal is $\tau \sim \mathrm{Exp}(\lambda)$, and we define $q = \mathbb{P}(\tau > t_{\max}) = e^{-\lambda t_{\max}}$ as the probability that a window ends without an event.

\begin{figure}[ht]
\centering
\begin{minipage}{0.48\columnwidth}
\centering
\includegraphics[width=\columnwidth]{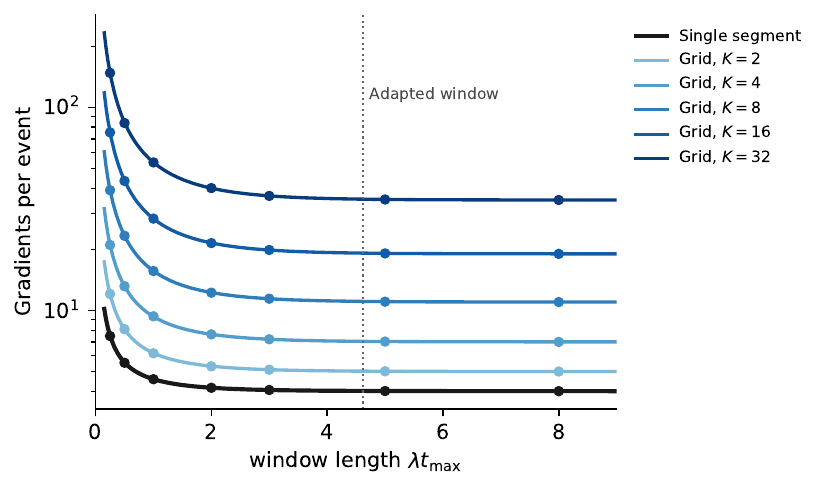}
\end{minipage}
\hfill
\begin{minipage}{0.48\columnwidth}
\centering
\includegraphics[width=\columnwidth]{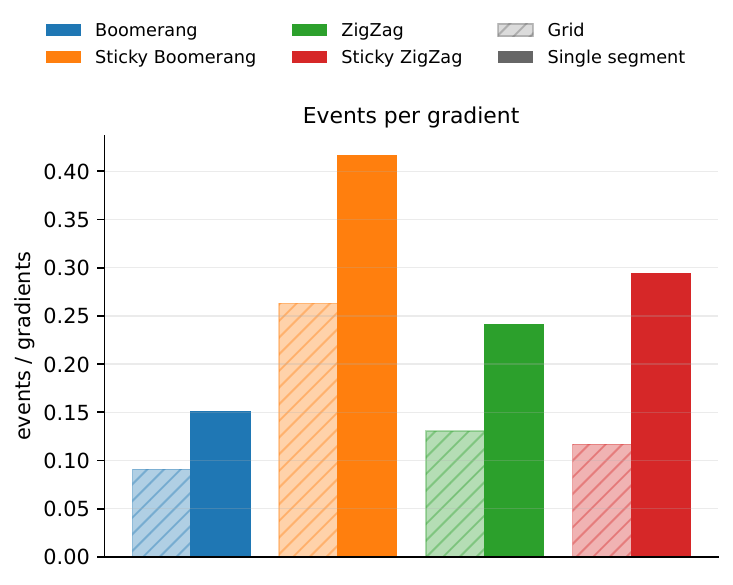}
\end{minipage}
\caption{Single segment against the grid bound. (a) Gradient evaluations per event under a constant rate $\lambda$ as a function of the window length, from the expressions above (lines) and a simulation of both schemes (dots), with the window the adaptation settles at dotted. (b) Events per gradient evaluation with the grid (hatched) and the single segment (solid), on the small network for a single split of the Boston data.}
\label{fig:ss_vs_grid}
\end{figure}

The cost of a window depends on whether it ends in an event, and on whether the previous window did. The grid evaluates its $K+1$ grid points and the single segment its $2$ end points. If the previous window was empty, its right end is the left end of the current window and can be reused, so the grid needs $K$ evaluations and the single segment $1$. On top of this, every proposal costs one evaluation of the rate, and an accepted proposal one more for the reflection. With $r$ rejected proposals in the window, the four cases are
\begin{center}
\begin{tabular}{llcc}
\toprule
Previous & Current & Single segment & Grid \\
\midrule
Empty ($q$) & Empty ($q$) & $1 + r$ & $K + r$ \\
Empty ($q$) & Event ($1-q$) & $3 + r$ & $K + 2 + r$ \\
Event ($1-q$) & Empty ($q$) & $2 + r$ & $K + 1 + r$ \\
Event ($1-q$) & Event ($1-q$) & $4 + r$ & $K + 3 + r$ \\
\bottomrule
\end{tabular}
\end{center}
Since the windows are independent under a constant rate and fixed $t_{\max}$, the expected cost of a window is
\begin{align}
    \mathbb{E}[G_{\text{ss}}] &= q\big(q + 3(1-q)\big) + (1-q)\big(2q + 4(1-q)\big) + \mathbb{E}[r] = 1 + 3(1 - q) + \mathbb{E}[r], \\
    \mathbb{E}[G_{\text{grid}}] &= q\big(Kq + (K+2)(1-q)\big) + (1-q)\big((K+1)q + (K+3)(1-q)\big) + \mathbb{E}[r] = K + 3(1 - q) + \mathbb{E}[r].
\end{align}
Rejections arrive at rate $\varepsilon\lambda$ along the distance travelled, which is $\tau$ if the window ends in an event and $t_{\max}$ otherwise, so
\begin{equation}
    \mathbb{E}[r] = \varepsilon\lambda\,\mathbb{E}[\tau \mid \tau < t_{\max}]\,(1-q) + \varepsilon\lambda\,t_{\max}\,q = \varepsilon\,(1-q).
\end{equation}
A window holds at most one event, so dividing by $1-q$ gives the expected cost per event,
\begin{equation}
    C_{\text{ss}} = \frac{1}{1-q} + 3 + \varepsilon,
    \qquad
    C_{\text{grid}} = \frac{K}{1-q} + 3 + \varepsilon .
\end{equation}
With the adaptation rule of \Cref{sec:supp_bounds}, both schemes settle where $q = \mathbb{E}[r]$, that is $q = \varepsilon/(1+\varepsilon)$ and $\lambda t_{\max} \approx 4.6$ for a constant rate. There
\begin{equation}
    C_{\text{ss}} = 4 + 2\varepsilon \approx 4,
    \qquad
    C_{\text{grid}} = K(1+\varepsilon) + 3 + \varepsilon \approx K + 3,
\end{equation}
so the single segment needs about $(K+3)/4$ times fewer gradient evaluations per event. Since empty windows are rare at the adapted window, the reuse of the end point matters little, and the gain comes from building each window from $2$ evaluations instead of $K+1$. For the sticky samplers, freezes and thaws are not proposed by thinning. Their times are known in advance, and the window simply ends there. Such windows are usually short and contain no proposal, so they only cost the construction of the bound, $2$ evaluations with a single segment and $K+1$ with the grid. For short windows the grid uses its minimum of $K = 2$ segments, so the gain is about $3/2$ per freeze or thaw. A simulation of both schemes agrees with these expressions (\Cref{fig:ss_vs_grid}, a).

On real data, we compare the two schemes on the small network for Boston split 0 at $w = 0.3$, where every run of both versions used exactly $10^6$ gradient evaluations (\Cref{fig:ss_vs_grid}, b).\footnote{Like Algorithm 4 of \citet{andral2024}, our grid runs did not reuse the end point of an empty window. Under the constant-rate model this raises their cost more where empty windows are frequent.} The single segment produces 1.84 (ZigZag), 2.52 (sticky ZigZag), 1.66 (Boomerang) and 1.58 (sticky Boomerang) times as many events per gradient evaluation.\footnote{Both implementations recompute the gradient for the bounce. Reusing the one from the accepted proposal would save one evaluation per bounce for both bounds, so the reported costs are conservative.} The gain for the sticky Boomerang, whose events are mostly freezes and thaws, is close to the $3/2$ predicted above for such events.

\subsection{More MNIST results}
\label{sec:supp_mnist}
\Cref{tab:MNIST} reports the predictive performance on the 8000 MNIST test images not used for pruning. The sticky samplers are sparser than $\beta_{\mathrm{ref}}^{\star}$, with 98\% against 91\% of the weights at zero for the FFN, and lose at most 0.7 percentage points of accuracy against SGD. The pruned MAP has the lowest NLL on the FFN. The dense ZigZag has the lowest accuracy but the lowest calibration error on the FFN, and the highest accuracy, second to lowest NLL and lowest Brier score on the LeNet-5. The sticky Boomerang is slightly behind the sticky ZigZag on every predictive metric.

\begin{figure}[t]
\centering
\begin{minipage}[c]{0.3\textwidth}
    \centering
    \includegraphics[width=\linewidth]{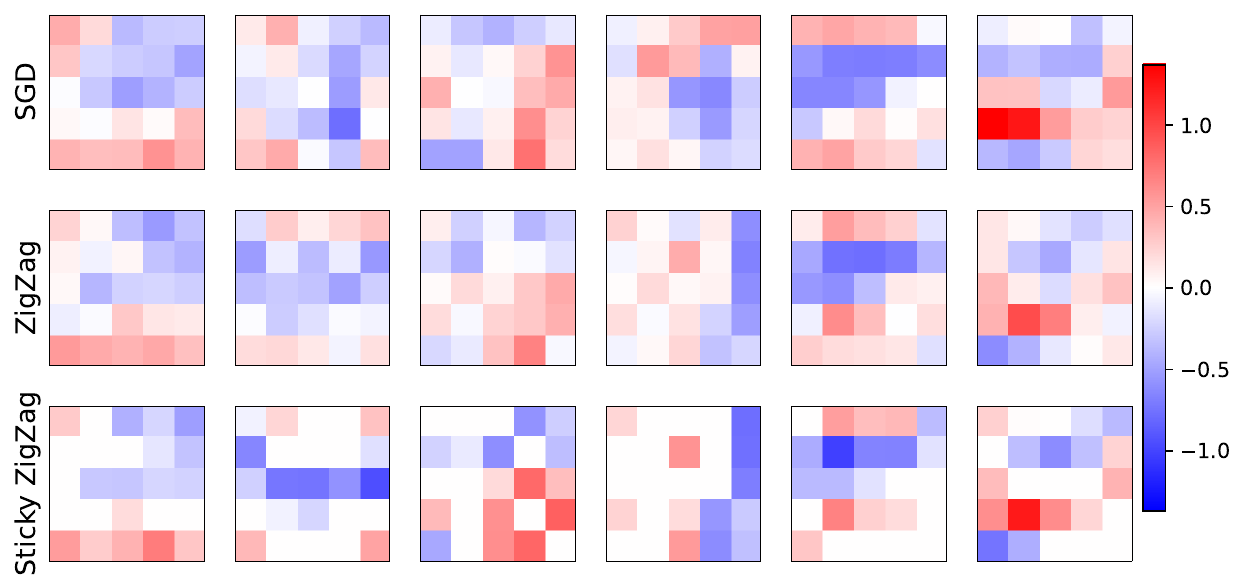}\\
    \small (a)
\end{minipage}
\hfill
\begin{minipage}[c]{0.66\textwidth}
    \centering
    \includegraphics[width=\linewidth]{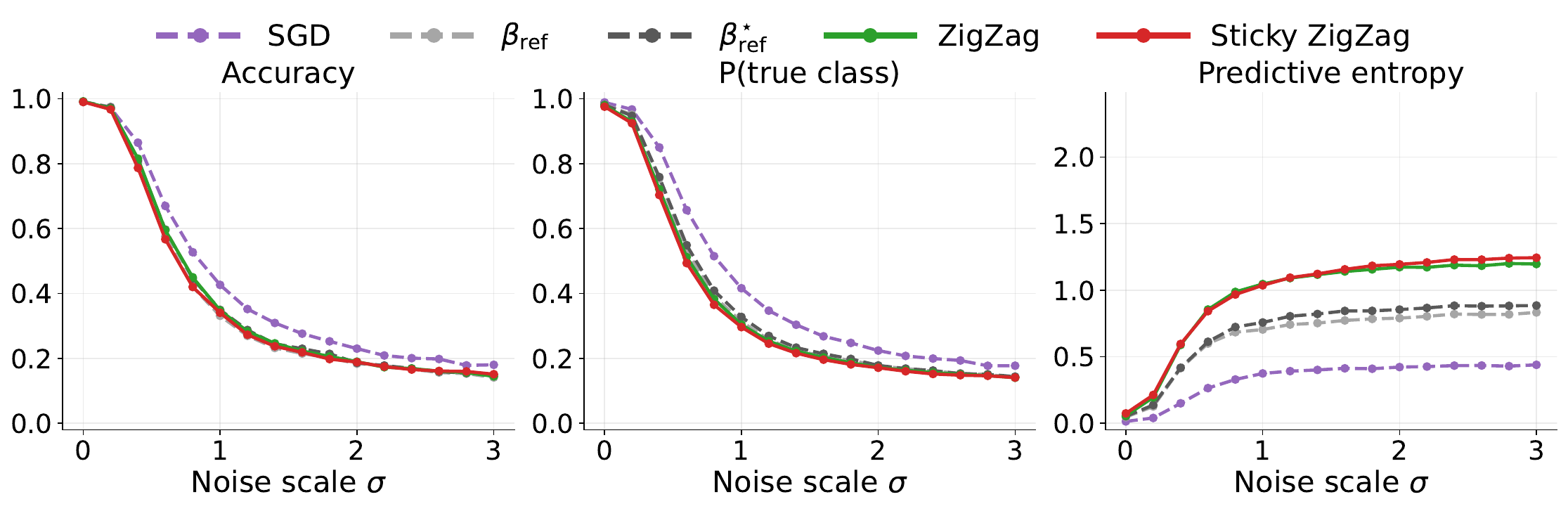}
    \small (b)
\end{minipage}
\caption{The analyses of \Cref{fig:MNIST_noise} for the LeNet-5. (a) The six first-layer filters. (b) Robustness to Gaussian noise of standard deviation $\sigma$ on the $[0,1]$ pixel scale.}
\label{fig:supp_lenet}
\end{figure}

The weight plots show the last draw of each sampler on a colour scale shared across rows. For the FFN, each row shows the four hidden units with the largest input-weight norm. Units are not matched across rows, as SGD is fitted separately. LeNet-5 has no weight per pixel, and its first layer consists of six $5\times5$ filters shared across all positions (\Cref{fig:supp_lenet}a). These filters are structured for both samplers, and the dense and sticky ZigZag are about equally robust to noise (\Cref{fig:supp_lenet}b).

\begin{table}
\centering
\resizebox{\columnwidth}{!}{
\begin{tabular}{l ccccc ccccc}
\toprule
& \multicolumn{5}{c}{FFN} & \multicolumn{5}{c}{LeNet-5} \\
\cmidrule(lr){2-6} \cmidrule(lr){7-11}
Model & Sparsity & Acc & ECE & NLL & Brier & Sparsity & Acc & ECE & NLL & Brier \\
\midrule
SGD                    & --   & \textbf{0.9836} & 0.0117 & 0.0854 & 0.0277 & --   & 0.9912 & 0.0053 & 0.0342 & 0.0148 \\
$\beta_{\mathrm{ref}}$ & --   & 0.9812 & 0.0070 & 0.0620 & 0.0284 & --   & 0.9908 & \textbf{0.0048} & \textbf{0.0297} & 0.0146 \\
ZigZag                 & --   & 0.9756 & \textbf{0.0065} & 0.0806 & 0.0370 & --   & \textbf{0.9923} & 0.0088 & 0.0307 & \textbf{0.0142} \\
$\beta_{\mathrm{ref}}^{\star}$ & 91\% & 0.9821 & 0.0073 & \textbf{0.0585} & \textbf{0.0273} & 89\% & 0.9909 & 0.0055 & 0.0310 & 0.0144 \\
Sticky ZigZag          & 98\% & 0.9797 & 0.0164 & 0.0759 & 0.0339 & 92\% & 0.9904 & 0.0107 & 0.0369 & 0.0163 \\
Sticky Boomerang       & 98\% & 0.9766 & 0.0210 & 0.0840 & 0.0373 & 94\% & 0.9883 & 0.0147 & 0.0447 & 0.0192 \\

\bottomrule
\end{tabular}
}
\caption{Predictive performance on the 8000 MNIST test images not used for pruning, for the FFN and LeNet-5.}
\label{tab:MNIST}
\end{table}

\subsection{Minibatching}
\label{sec:supp_minibatch}
One of the most attractive properties of PDMPs is that they admit exact subsampling. Assuming i.i.d. data, the full-data gradient $\nabla U$ can be replaced by an unbiased estimate based on a single randomly chosen observation without changing the invariant distribution, provided the rate bound dominates the switching rate of every observation-specific gradient estimate along the trajectory \citep{bierkens2019bigdata}. For BNNs, to our knowledge, no such bound is available, and constructing one numerically by evaluating every per-observation rate over the bounding horizon is at least as costly as a full-batch pass. \citet{corbella2022} instead estimate the population bound statistically, fitting a generalized Pareto distribution to the maxima of a subset of observation-specific rates and extrapolating to the largest one. The resulting bound is not guaranteed to dominate every rate, and extending such schemes to high-dimensional BNNs is an interesting direction for future work. For this paper however, we use an approximate scheme. Each time the rate bound is constructed, we draw a fresh minibatch of observations and use it for all gradient evaluations in that bounding step, both for the bound and for the accept–reject decisions. Within one bounding horizon, the sampler therefore simulates the PDMP targeting the rescaled minibatch posterior, and a new minibatch is drawn for each horizon. The noise in the minibatch gradients introduces bias, so the process no longer has the exact posterior as its invariant distribution. We apply minibatching to our MNIST models.

To assess the effect of this approximation, \Cref{tab:mb} compares minibatched and full-batch runs of the small BNN on the UCI datasets. Minibatching does not change predictive performance systematically. It is worse in some cases, most notably the Yacht NLL of both ZigZag samplers and the Boomerang on Boston, and better in others, most notably the Sticky ZigZag on Concrete. While this does not guarantee the same behaviour for the MNIST models, where a full-batch comparison is infeasible within our budget, it suggests that the approximation has a limited effect on predictive performance.
\begin{table}[t]
\centering
\small
\setlength{\tabcolsep}{3pt}
\begin{tabular}{l rrr rrr rrr rrr}
\toprule
 & \multicolumn{3}{c}{Boston} & \multicolumn{3}{c}{Energy}
 & \multicolumn{3}{c}{Concrete} & \multicolumn{3}{c}{Yacht} \\
\cmidrule(lr){2-4} \cmidrule(lr){5-7} \cmidrule(lr){8-10} \cmidrule(lr){11-13}
$\Delta$ & RMSE & NLL & CRPS & RMSE & NLL & CRPS & RMSE & NLL & CRPS & RMSE & NLL & CRPS \\
\midrule
Boomerang        & -.08 & -.08 & -.11 & -.00 & -.03 & -.00 & -.01 & -.03 & -.06 & .01 & .04 & .04 \\
ZigZag           & -.11 & -.08 & -.07 & -.00 & -.08 & -.00 & .09 & -.07 & .09 & -.01 & -.31 & -.01 \\
Sticky Boomerang & -.02 & -.01 & .01 & -.00 & -.00 & -.00 & .08 & .03 & .06 & -.00 & .19 & -.00 \\
Sticky ZigZag    & .04 & .00 & .03 & -.01 & -.07 & -.00 & .42 & -.00 & .27 & -.01 & -.69 & -.01 \\
\bottomrule
\end{tabular}
\caption{Difference in predictive performance between full-batch and minibatch runs of the small network, averaged over the 5 splits of each UCI dataset. Values are full batch minus minibatch on the original scale, so negative values indicate a performance loss from minibatching. The minibatch runs use batches of 32 observations and $10^5$ events, the full-batch runs $10^6$ gradient evaluations. Both use the grid bound of an earlier version of our implementation.}

\label{tab:mb}
\end{table}

\end{document}